\documentclass[twoside,11pt]{article}

\usepackage{jmlr2e}

\usepackage{epsfig}
\usepackage{amssymb} 
\usepackage[tbtags]{amsmath} 
\usepackage{graphics,eepic,epic}
\usepackage{latexsym}
\usepackage{euscript}
\usepackage{graphics,eepic,epic,psfrag}
\usepackage{color}
\usepackage{algorithmic}
\usepackage{algorithm}
\usepackage{subcaption}
\usepackage{hyperref}
\usepackage{array}
\usepackage{tikz}
\usetikzlibrary{calc}

\DeclareMathOperator*{\argmax}{arg\,max}
\DeclareMathOperator*{\argmin}{arg\,min}

\renewcommand{\cite}{\citep}

\DeclareMathOperator*{\subjectto}{subject\,to}
\newcommand{\st}{\text{ s.t. }}

\newcommand{\varSet}{\mathcal{I}_v}
\newcommand{\varSetInd}[1]{\mathcal{I}_{v,#1}}
\newcommand{\facSet}{\mathcal{I}_f}
\newcommand{\mfacSet}{\mathcal{I}_m}
\newcommand{\efacSet}{\mathcal{I}_e}
\newcommand{\abstSet}{\mathcal{A}}
\newcommand{\hood}[1]{\mathcal{N}_{#1}}
\newcommand{\edgeSet}{\mathcal{E}}
\newcommand{\edge}[2]{e_{#1}^{#2}}
\newcommand{\vote}[2]{v_{#1}^{#2}}
\newcommand{\voteVec}[1]{v^{#1}}

\newcommand{\opin}[2]{o_{#1}^{#2}}
\newcommand{\opinVec}[1]{o^{#1}}
\newcommand{\msg}[2]{m_{#1}^{#2}}
\newcommand{\reactiveCost}[1]{\phi_{#1}}
\newcommand{\summFacSet}[2]{\mathcal{N}_{#1}^{\backslash#2}}

\newcommand{\optVar}[1]{\tilde{x}_{#1}}

\newcommand{\wt}[2]{w_{#1}^{#2}}

\newcommand{\wtSum}[1]{\hat{w}_{#1}}

\newcommand{\locCost}[1]{\psi_{#1}}
\newcommand{\selCost}[1]{\chi_{#1}}

\newcommand{\changeSet}{\mathcal{V}}
\newcommand{\reactSet}{\mathcal{R}}
\newcommand{\dissSet}{\mathcal{D}}

\newcommand{\constSet}{\mathcal{I}_c}

\newcommand{\mem}[2]{\mu^{#1}_{#2}}
\newcommand{\memInd}[3]{\mu^{#1}_{#2,#3}}

\newcommand{\ind}{\mathbb{I}}

\newcommand{\tcb}{\textcolor{blue}}

\ShortHeadings{Proactive Message Passing on Memory Factor Networks}{Eschenfeldt, Schmidt, Draper, and Yedidia}
\firstpageno{1}

\begin{document}

\title{Proactive Message Passing\\ on Memory Factor Networks}

\author{\name Patrick Eschenfeldt \email peschen@mit.edu \\
       \addr Operations Research Center\\
       Massachusetts Institute of Technology\\
       Cambridge, MA 02139, USA
       \AND
       \name Dan Schmidt \thanks{As of January 2016 he is with Analog Devices Lyric Labs, Cambridge, MA 02142.} \email dan.schmidt@disneyresearch.com \\
       \addr Disney Research Boston\\
       Cambridge, MA 02142, USA 
       \AND
       \name Stark Draper \email stark.draper@utoronto.ca \\
       \addr Dept.~of Electrical and Computer Engineering\\ 
       University of Toronto, ON M5S 3G4, Canada
       \AND
       \name Jonathan Yedidia \footnotemark[1] \email yedidia@disneyresearch.com \\
       \addr Disney Research Boston\\
       Cambridge, MA 02142, USA }

\editor{TBD}

\maketitle

\begin{abstract}%
  We introduce a new type of graphical model that we call a 
  ``memory factor network''
  (MFN).  We show how to use MFNs to model the structure inherent in 
  many types of data sets.  We also introduce an associated
  message-passing style algorithm called ``proactive message passing'' (PMP)
  that performs inference on MFNs. PMP comes with
  convergence guarantees and is efficient in comparison to
  competing algorithms such as variants of belief propagation.  We
  specialize MFNs and PMP to a number of distinct types of data
  (discrete, continuous, labelled) and inference problems
  (interpolation, hypothesis testing), provide examples, and discuss
  approaches for efficient implementation.
\end{abstract}


\section{Introduction and motivations}
\label{sec.intro}

In this paper, we introduce ``memory factor networks'' (MFNs) and the
``proactive message passing'' (PMP) algorithm.  Our objective is to
combine the capability of message passing algorithms, which can make
large-scale inferences in a highly efficient manner, with the ability
of machine learning algorithms to generalize from experience.


Factor graphs and message
passing~\cite{kschischangEtAl,loeligerEtAl,koller,YedidiaBP,sudderth} have proved to be an
extremely effective combination when one is faced with an
inference task wherein global problem structure decomposes
into a large set of local constraints.  In applications such as error
correction decoding \cite{richardsonUrbanke}, these local constraints have a simple
characterization that makes local inference computationally simple.
Message-passing algorithms iterate between making local inferences and
combining these local decisions into a global estimate.

However, in many settings the local problem structure is not known ahead
of time and engineered, as it is in error correcting codes, but rather
must be learned.  For instance, a number of examples may be provided
and the initial task is to deduce underlying problem structure from these
exemplars.  Following that, one can make inferences based on the
problem structure learned.  These tasks---of deducing problem structure
and then making inferences based on the structure---are 
central to many areas of data
analysis, statistics, and machine learning.

In MFNs a set of variables describe a problem configuration.  A
collection of ``memory factors'', learned from examples, encode
(possibly soft) constraints on overlapping subsets of the variables.
The high-level goal of PMP is to find a configuration of variables
that is minimally conflicting with the local problem structure
represented by the memory factors and with the evidence obtained from
the world.  Conflicts occur because the constraints overlap and, in
general, cannot all be satisfied exactly.

While the above description is somewhat generic, one key innovation in
MFNs and PMP is to design the memory factors so that they encode the
learned local problem structure in a manner easily accessible for
inference.  The guiding philosophy is one of joint design of the
graphical model and of the message passing algorithm so as to make the
necessary computations efficient.  In analogy with biology, we need our
memories to be easily accessible to our thinking.  Thus, while we
do not engineer the detailed statistical 
structure of the problem, what we do engineer is
the way in which we store our knowledge of that statistical structure in order to
make it easy to exploit in our inference tasks.  In this framework,
``memories'' correspond to the learned local problem structure
encoded by memory factors.

A second key innovation in PMP is a simple methodology for producing
local inferences, one that guarantees global convergence.  Because the
learned problem structure encoded into the memory factors is extremely
rich---in comparison to the factor nodes used in graphical
models which belief
propagation is typically applied to---a 
simple methodology for producing local inference is
important to maintain computational tractability.  PMP accomplishes
this through an intuitive voting mechanism that, at each iteration,
considers the current messages coming from neighboring memory factors
in order only to make choices that reduce a global objective.  PMP is
thus ``proactive'' in the sense that it considers the effect that each
possible local choice will have on the global situation before deciding
which choice to implement.  This proactivity ensures that the global
objective decreases at each iteration, and guarantees that PMP will
converge to a locally optimum configuration.  In order to ensure that
PMP converges to a good local optimum, we use a powerful heuristic for
scheduling factor updates, one that is derived from a novel notion of
factor ``confidence.''

The general idea of combining a message-passing inference algorithm with a graphical
model built up using an example-based learning approach was introduced
by Freeman et al. in their VISTA (``Vision by Image/Scene training'') approach
\cite{learningLowLevel}. The main application considered using 
the VISTA approach was
example-based super-resolution \cite{superResolution}.
The VISTA approach used Markov random field graphical models, built up dynamically in 
response to each inference problem encountered (that is, only the most
probable example patches were included in each Markov random field node) and
the inference algorithms used were belief propagation or more efficient
one-pass algorithms. One can think of this paper as providing a 
formal generalization of the
VISTA idea. That is, whereas the VISTA work was primarily concerned
with finding a good solution to the particular application of super-resolution, our primary
concern is to carefully define and demonstrate a general approach to learning-based inference, based
on the memory factor network data structure and the proactive message-passing algorithm. We believe 
this approach
can be applied to a wide range of problems going beyond those encountered in computer vision.

The rest of the paper is structured as follows.  In
Section~\ref{sec.MFN} we introduce memory factor networks.  In
Section~\ref{sec.PMP} we describe the proactive message passing
algorithm.  Then, in Section~\ref{sec.varTypes} we specialize the
discussions of Section~\ref{sec.MFN} and~\ref{sec.PMP} to specific
types of variable nodes (integer, real, label) and cost functions
(absolute difference, quadratic difference, indicator).  In
Section~\ref{sec.memFacTypes} we turn to memory factors and describe
two distinct types of memory factors.  We first consider ``memory
tables'' where each memory factor consists of a database of fragments of observed
exemplars. We then consider ``subspace factors''
where the memory factor constrains local problem structure to reside
in some learned low-dimensional subspace.  In
Section~\ref{sec.applications} we provide illustrative applications
that include face reconstruction from missing and noisy data, music
reconstruction, and handwritten digit classification.  In some
examples we compare and contrast the performance of memory tables with
subspace factors.  We make concluding remarks in
Section~\ref{sec.conclusion}.

\section{Memory factor networks}
\label{sec.MFN}

A memory factor network (MFN) contains $N$ variables and $M$
constraints on (subsets of) those variables.  Constraints may encode
problem structure or may be induced by observations.  We capture these
two types of constraints through ``memory factor'' and ``evidence
factor'' nodes respectively.  We seek a configuration of the $N$
variables that minimizes a cost function of the variables induced by
the $M$ factors.  In the following we use the notation $\{1, \ldots,
N\} = [N]$ to denote an index set of cardinality $N$.

The interconnection of variables and factors is described by an
edge-weighted bipartite graph.  That graph, and the cost function formulation, is
described by the following sets:
\begin{itemize}
\item A set of variable nodes indexed by $\varSet = [N]$ with values
  $\{x_i : i \in \varSet\}$.  We will discuss various choices for the
  alphabet $\mathcal{X}_i$ of $x_i$ --- real, integer, binary, label.
  Distinct variables can have different alphabets.
\item A set of factor nodes indexed by $\facSet = [M]$.  As mentioned,
  factors are either memory factors, indexed by $\mfacSet$, or
  evidence factors, indexed by $\efacSet$, where $\mfacSet \cup
  \efacSet = \facSet$ and $\mfacSet \cap \efacSet =
  \emptyset$. 
\item A set of edges $\edgeSet = \{\edge{i}{a} : i \in \varSet, a \in
  \hood{i}\}$ induced by the neighborhood structure of the network.
  With slight (but always resolvable from context) equivocation in
  notation we write $\hood{i}$ ($\hood{a}$) to denote the set of
  factor (variable) nodes neighboring variable node $i$ (factor node
  $a$).  Note that each evidence factor is typically attached to a single
  variable, so that if $a \in \efacSet$, then typically $|\hood{a}| = 1$.
\item A set of votes $\{\vote{i}{a} : i \in \varSet, a \in \hood{i}\}$
  and corresponding vote weights $\{\wt{i}{a} : i \in \varSet, a \in
  \hood{i}\}$.  We always have $\vote{i}{a} \in
  \mathcal{X}_i$. The weights ${\wt{i}{a}}$ are fixed real numbers, and we normally restrict
  them so that all weights associated with the same variable $i$ are equal.
\end{itemize}

When an MFN is being used for inference, the degrees of freedom in a
memory factor network are the values of the variable nodes $\{x_i : i
\in \varSet\}$ and the votes of the factor nodes $\{\vote{i}{a} : i
\in \varSet, a \in \hood{i}\}$. Changes in the state of the world are
reflected in changes to the evidence factors.  When an MFN is
learning, the acquisition of memories is reflected in adjustments to
the memory factors which, at inference time, is reflected in the
factor's possible votes.

We often have reason to refer to the set of variables or votes or
weights associated with a single factor.  We use $x^a = \{x_i : i \in
\hood{a}\}$ to refer to the vector of variables neighboring factor $a$
and use $\voteVec{a}$ and $w^a$, similarly defined, to denote the
``vote'' vector and the ``weight'' vector.  Occasionally we have need
to refer to, say, the set of variables in some set $\mathcal{S}
\subset \varSet$; we write $x_{\mathcal{S}}$ to refer to this set.  We
also will have need to refer to a set of votes regarding variable $i$
from some subset $\mathcal{B} \subseteq \hood{i}$ of neighbors; we
denote this set $\vote{i}{\mathcal{B}}$.

Perhaps other than the edge weights, the above description is
standard; it quite neatly fits into the formalism of a ``normal''
factor graph due to Forney~\cite{forney}.  To map to Forney's
normal representation, the variable and factor nodes would all
correspond to factors (the former to equality constraints), and votes
would correspond to variables.

Our objective is to minimize a global cost function $\Psi$ that can be
factored based on the graphical structure of the problem. 
$\Psi$ will include two types of local costs:
mismatch costs and selection costs. The mismatch costs penalize the
difference between a variable setting $x_i$ and a factor's vote on
that variable $\vote{i}{a}$.  The contribution of mismatches to the
global cost is additive in both variable and factor indices, and each
contribution is weighted by the particular vote's weight. We make the
mismatch costs a function of the variable index, 
but not the factor index.  The reason is that, other
than the weighting, all votes pertaining to a particular variable are
compatible in some sense (e.g., all in $\mathbb{R}$ or in
$\mathbb{Z}$) and therefore the same measure of mismatch should apply
to all.  The selection costs make various choices of a factor's vote
vector $\voteVec{a}$ more or less costly and thus is a function of the
pertinent factor's index $a$.  Unlike the mismatch costs, selection costs,
are, in
general, only additive in factor indices and not in variable indices.
To denote the mismatch costs we use $\locCost{i}(\cdot, \cdot)$; to
denote selection costs we use $\selCost{a}(\cdot)$.
\begin{align}
  \Psi & = \sum_{a \in \facSet} \left[ \selCost{a}(\voteVec{a}) +
    \sum_{i \in \hood{a}} \locCost{i}(x_i, \vote{i}{a}) \wt{i}{a}
  \right]
  \label{eq.globalCost}
\end{align}
As we detail more in later sections, we consider various forms for
both types of cost functions.  For example, for local mismatch costs
we may use the absolute difference $\locCost{i}(u,v) = |u-v|$ when
$u,v \in \mathbb{Z}$, the squared difference $\locCost{i}(u,v) =
|u-v|^2$ when $u,v \in \mathbb{R}$, and an indicator function
$\locCost{i}(u,v) = \ind[u \neq v]$ where $u,v$ are in a discrete set.
For selection costs, we can design $\selCost{a}(\cdot)$ to limit
$\{\vote{i}{a}\}_{i \in \hood{a}}$ to belong to some discrete set or
to lie in a certain subspace of a vector space.

In practice, the selection costs arise as a result of a training phase
where examples are shown to the system and likely or frequent
configurations are learned.  Later, when the system is trying to
understand a new sample from the world, it tries to minimize the
overall cost, thus interpreting the new sample in terms of both how well
the variables can be made to match the evidence from that sample 
and how well the variables can match the structure learned from training examples.




\section{Proactive message passing}
\label{sec.PMP}

In our efforts to minimize~(\ref{eq.globalCost}) we now introduce the
proactive message passing (PMP) algorithm.  Proactive message passing
works through a sequence of iterative subproblems in its attempt to
minimize~(\ref{eq.globalCost}).  The PMP algorithm can be understood
through analogy to an idealized political convention, where the objective of
the convention is to determine the party platform.  A party platform
consists of a number of stances on issues, where each issue corresponds to one
variable in the analogy, and a stance is a value for the variable. 
Each delegate corresponds to one factor 
and each delegate is
concerned with only a (generally small) subset of issues.  By the
close of the convention we want to identify the optimal vector of
stances that results in minimal dissatisfaction amongst delegates;
dissatisfaction is measured by~(\ref{eq.globalCost}).  While the
analogy with a political convention is not exact, we hope it will help
elucidate the PMP algorithm and develop intuition.


In a convention there is a sequence of rounds of balloting.  Each
round corresponds to one iteration of the algorithm.  In each round
some delegates vote on the issues of concern to them, while others
abstain from voting.  The delegates that vote initially are the ones
most confident in their opinions.  Other delegates temporarily abstain
and delay voting until they become sufficiently confident in their
opinions.  (In PMP, once a factor votes, it votes on all variables of
interest to it, and never abstains again.)  After each round of
balloting, all delegates reconsider their opinions, which may have
changed based on the most recent set of votes cast.  When forming
opinions, a delegate takes into account all current votes (of
non-abstaining delegates) on each issue that is of interest to that
delegate.  The opinions formed correspond to votes that, if cast,
would reduce maximally a cost function derived
from~(\ref{eq.globalCost}).  This is why the algorithm is termed
``proactive,'' as each delegate preemptively evaluates how a change in
their votes would impact the dissatisfaction of all delegates that
have voted on the same set of issues.  In the next round of balloting
some subset of delegates is chosen to change their votes or to cast
their initial set of votes.  (Delegates that are not chosen to cast
new votes, but that have previously voted, leave their votes
unchanged.)
This subset is chosen to be the set of delegates most confident in
their opinions; that is, they are the delegates least likely to have
their opinions change in the future due to other delegates' votes.
The cost function is structured so that, eventually, all delegates
must cast votes and not abstain.  The convention ends when all
delegates' votes align sufficiently with their opinions that no one
wants to change their votes.

%
%

\subsection{PMP specification}

We now put a mathematical framework around the intuition described
above.  As in the above description, each factor has a
current ``opinion'' on the value of each variable neighboring it, 
an opinion that may differ from a
vote it has already cast.  The opinion vector is in one-to-one
correspondence with the vote vector, and is denoted as
$\{\opin{i}{a} : i \in \varSet, a \in \hood{i}\}$.

The iterative subproblems that PMP works through correspond to
the rounds of balloting.  In each iteration some subset of factors
$\abstSet \subseteq \facSet$ ``abstain'' from voting.  One part of the
objective of each subproblem is a modification of the original
objective that considers only the participating (non-abstaining)
factors:
\begin{align}
  \Psi_{\abstSet} = \sum_{a \in \facSet \backslash \abstSet} \left[
    \selCost{a}(\voteVec{a}) + \sum_{i \in \hood{a}} \locCost{i}(x_i,
    \vote{i}{a}) \wt{i}{a} \right].
  \label{eq.actSetCost}
\end{align}
To connect to the original objective~(\ref{eq.globalCost}), PMP 
minimizes a cost tuple $(|\abstSet|, \Psi_{\abstSet})$.  The first element
of the tuple is the count of abstaining factors, and the second
is~(\ref{eq.actSetCost}).  To compare tuples, we assert a
lexicographic order: $(a_1, b_1) \leq (a_2, b_2)$ if and only if
either $a_1 < a_2$ or $a_1 = a_2$ and $b_1 \leq b_2$.  With this
ordering the global optimum of the tuple is $(0, \Psi^{\ast}$) where
$\Psi^{\ast}$ minimizes~(\ref{eq.globalCost}). Our PMP algorithm will have
the feature that at each iteration, the cost will decrease, and ultimately
no factors will abstain, although the algorithm may converge to a local
optimum rather than the global one. This feature of guaranteed convergence
to at least a local optimum is not shared by most message-passing algorithms
operating on factor graphs containing cycles~\cite{YedidiaBP}.

In addition to the abstaining set, PMP works with three other sets of
factors: the vote-changing set $\changeSet$, the reacting set
$\reactSet$ and the dissatisfied set $\dissSet$.  At each iteration
(ballot) there is a set of factors (delegates) that change their
votes.  The set of these factors is the vote-changing set
$\changeSet$.  If a factor is in $\changeSet$ and has not voted
previously it is removed from the abstaining set $\abstSet$.  Note
that we never add factors to $\abstSet$ and therefore the cardinality
of the abstaining set $|\abstSet|$ decreases monotonically in
iteration count.  The reacting set $\reactSet$ is the set of factors
that neighbor variables connected to the set of factors that most recently changed
their votes:
$\reactSet = \cup_{a \in \changeSet} (\cup_{i \in \hood{a}} \hood{i}
\backslash \{a\})$.
This is the set of factors that, based on the most recent change in
votes, might change their opinions about some variables (issues).  The
dissatisfied set $\dissSet$ contains factors whose recomputed opinion
vector does not in fact match their vote vector but have not yet been
given a chance to cast votes that match their latest opinions.  Note
that while the abstaining set plays a fundamental role in the
algorithm, the role of the reactive and dissatisfied sets is purely to
reduce computation. The algorithm could be defined without tracking
these latter two sets, but a
number of redundant computations would then be repeated each
iteration.

To initialize the algorithm we start with a set of factors
$\changeSet$ that we want to vote in the first iteration.  One natural
choice is to set $\changeSet = \efacSet$, the set of evidence nodes,
and to set the votes of these factors equal to their observations.
Then we set the abstaining set
$\abstSet = \facSet \backslash \changeSet$, the reacting set
$\reactSet = \cup_{a \in \changeSet} (\cup_{i \in \hood{a}} \hood{i}
\backslash \{a\})$, and the dissatisfied set $\dissSet = \emptyset$.

We note that we use selection and mismatch costs, $\selCost{a}(\cdot)$
and $\locCost{i}(\cdot, \cdot)$, that are non-negative.  Therefore,
both elements of the cost tuple associated with PMP are non-negative.
If, in a given iteration, any factor node stops abstaining,
then the first element of the tuple decreases, reducing the cost.  If
on the other hand the abstaining set does not change, then only the
second element of the tuple can change.  Since, by design, any choice
of new votes can only decrease~(\ref{eq.actSetCost}), then, when the
abstaining set does not change, the second element of the tuple is at
least as small as in the previous iteration.  Thus, the cost tuple is
monotonically decreasing.

The overall proactive message passing algorithm is summarized in
Algorithm 1. The key step in the algorithm is when the optimal opinion
vector for each factor is computed using~(\ref{alg.costComp}). In this
step an opinion vector $o^a$ is chosen for factor $a$ to minimize the
sum of the selection cost for this factor $\selCost{a}(\bar{o})$, the
mismatch costs $\sum_{i \in \hood{a}} \locCost{i}(x_i, \bar{o}_i)
\wt{i}{a}$ for this factor, and the mismatch costs
\begin{equation*}
\sum_{i \in \hood{a}} \sum_{b \in (\hood{i}\backslash\{a\}) \backslash
  \abstSet} \locCost{i}(x_i, \vote{i}{b}) \wt{i}{b}
\end{equation*}
for the other factors that neighbor the same set of variables that
this factor neighbors.  We ``proactively'' choose the value of
neighboring variables $x_i$ that will optimize the combined costs of the second
and third terms, given the current state of the other factors' votes
$\vote{i}{b}$. This ``proactivity'' is reminiscent of the
``cavity method'' introduced in statistical mechanics for the study of
spin glasses~\cite{spinglass}.

Another step that should be highlighted is when we choose the most
{\em confident} factor(s) to change their votes. We do not simply want to greedily
choose the factor that can reduce the overall cost by the most, because there
might be two very different choices for the values of variables connected to
that factor that both reduce the overall cost
by a similar large amount, and committing to one of them prematurely could 
lead to convergence to
a poor local optimum. In particular, this can easily happen for factors that are
initially isolated from any other factors, including evidence nodes. They could reduce
the overall cost significantly by selecting any ``memory'' (set of variable values consistent with any
one of the examples learned during training), but voting too early in this way would just
lead to premature commitments leading ultimately to convergence to poor local optima. 
Instead, we want to prioritize factors for which one choice
of variable values is significantly better than any other choice---what we call
``confident'' factors. Such factors typically share variables with several other factors that
are already voting, and given those votes, one choice of a memory would reduce the overall
cost much more than any other.

%

\begin{algorithm}
\caption{Proactive message passing}
\begin{algorithmic}[1]
\WHILE{the vote changing set $\changeSet \neq \emptyset$}

\FOR{each factor in the reacting set $a \in \reactSet$}


\STATE{Compute the opinion vector $\opinVec{a}$ of factor $a$. The opinion 
    vector is
  chosen to minimize a combination of selection cost
  $\selCost{a}(\cdot)$ of the opinion vector and the aggregate
  dissatisfaction induced by that choice.  Dissatisfaction is measured
  as the sum of mismatch costs $\locCost{i}(\cdot, \cdot)$,
  appropriately weighted, computed with respect to the optimal choice
  of variable settings for the vector $x^a$ for each possible opinion vector.  
  In the following
  $\bar{o}$ is a vector of the same dimension and alphabet as
  $\voteVec{a}$.
\begin{equation}
  \opinVec{a}  = \argmin_{\bar{o}} \left\{ \selCost{a}(\bar{o}) + 
  \sum_{i \in \hood{a}} \min_{x_i} \left[ \locCost{i}(x_i, \bar{o}_i) \wt{i}{a} + 
    \sum_{b \in (\hood{i}\backslash\{a\}) \backslash \abstSet} \locCost{i}(x_i, \vote{i}{b}) \wt{i}{b}
  \right]\right\}. \label{alg.costComp}
\end{equation}
If there is a set of optimizers, one of which matches the previous
vote vector $\voteVec{a}$ , set $\opinVec{a} = \voteVec{a}$.}

\IF{$\opinVec{a} \simeq \voteVec{a}$} 

\STATE{Factor $a$'s opinion vector is approximately the same as its current
  vote vector and the factor is satisfied.  Set $\dissSet = \dissSet
  \backslash \{a\}$. (Note the specification of $``\simeq''$ is
  problem specific.)}

\ELSE \STATE{Factor $a$ is dissatisfied.  Set $\dissSet = \dissSet
  \cup \{a\}$.}  

\STATE{Compute the confidence $\kappa(a)$ of factor $a$.  (The
  computation of $\kappa(a)$ is problem specific.)}

\ENDIF

\ENDFOR

\STATE{Identify the factor(s) that will change their votes. A sequential update rule would
    update the single most confident factor using $\changeSet =
  \argmax_{a \in \dissSet} \kappa(a)$ where ties are broken at random.
  Simultaneous voting 
  rules are also possible, where $|\changeSet| > 1$, for example updating a fixed fraction of
  the most confident factors.}
 
\FOR{each factor $a \in \changeSet$}

\STATE{Set the factor's vote vector equal to its opinion vector: $\voteVec{a} = \opinVec{a}$.}

\ENDFOR

\STATE{Recompute the abstaining set by removing new voters: $\abstSet
  = \abstSet \backslash \changeSet$.}

\STATE{Recompute the dissatisfied set by removing factors that just
  changed their votes: $\dissSet = \dissSet \backslash \changeSet$.}

\STATE{Determine the reacting set of factors whose opinion vectors may change
  due to the most recent update in votes: $\reactSet = \cup_{a \in
    \changeSet} (\cup_{i \in \hood{a}} \hood{i} \backslash \{a\})$.}

\ENDWHILE

\STATE{Define the output set $\mathcal{O} = \{i | i \in \cup_{a \in
    \facSet \backslash \abstSet} \hood{a}$\}.  (For most problems, at
  this stage $\abstSet$ will always equal $\emptyset$ and $\mathcal{O}$
  will equal $\varSet$.)}

\STATE{Compute the optimal set of variable settings given the votes
  $\tilde{x}_{\mathcal{O}} = \arg \min_{x_{\mathcal{O}}} \sum_{a \in
    \facSet \backslash \abstSet} [\selCost{a}(\voteVec{a}) + \sum_{i
      \in \hood{a}} \locCost{i}(x_i, \vote{i}{a}) \wt{i}{a}]$.}

\STATE{{\bf Return} $\tilde{x}_{\mathcal{O}}$.  For variables not in
  the output set, $\{x_i | i \in \varSet \backslash \mathcal{O}\}$, return an
  ``unknown'' flag.}
\end{algorithmic}
\end{algorithm}

\subsection{Message-passing formulation}
\label{sec.PMPasMsgPass}

Equation~(\ref{alg.costComp}) is written in general form, but we want to manipulate
it into a form that reveals a message passing interpretation. In particular, we want
to obtain ``messages'' to a factor node from each of the variable nodes connected to it, informing
the factor about the preferences of the rest of the network for each variable node, in
the form of some sufficient statistic.
To that end, we begin by defining
$\tilde{x}_i$ to be the optimal choice of $x_i$ based on the votes of
variable $i$'s non-abstaining neighbors other than factor $a$, i.e.,
\begin{align}
  \tilde{x}_i = & \arg \min_{x_i} \left[ \sum_{b \in
      \summFacSet{i}{a}} \locCost{i}(x_i, \vote{i}{b}) \wt{i}{b}
    \right]. \label{eq.prototypeMin}
\end{align}
where
\begin{equation}
\summFacSet{i}{a} = (\hood{i}\backslash\{a\}) \backslash \abstSet \label{eq.summFacSet}
\end{equation}
is defined to be the set of non-abstaining memory factors neighboring
variable $i$, excluding factor $a$.

We next rewrite the  innermost argument in~(\ref{alg.costComp}) as
\begin{align}
  & \min_{x_i} \left[ \locCost{i}(x_i, \bar{o}_i) \wt{i}{a} + \sum_{b
      \in \summFacSet{i}{a}}
    [\locCost{i}(x_i, \vote{i}{b}) - \locCost{i}(\tilde{x}_i, \vote{i}{b}) + \locCost{i}(\tilde{x}_i, \vote{i}{b})] \wt{i}{b} \right]\\ 
= & \min_{x_i} \left[ \locCost{i}(x_i, \bar{o}_i) \wt{i}{a} + \sum_{b
      \in \summFacSet{i}{a}}
    [\locCost{i}(x_i, \vote{i}{b}) - \locCost{i}(\tilde{x}_i, \vote{i}{b})] \wt{i}{b}  + \sum_{b
\in  \summFacSet{i}{a}}  \locCost{i}(\tilde{x}_i, \vote{i}{b}) \wt{i}{b} \right], \label{eq.addSubtract}
\end{align}
and substitute the result back into~(\ref{alg.costComp}).  We can drop the
last term in (\ref{eq.addSubtract}) because it is not a function of either $x_i$ or $\bar{o}$,
yielding the following modified form of the optimization stated
in~(\ref{alg.costComp}):
\begin{align}
\opinVec{a}  = \argmin_{\bar{o}} \left\{ \selCost{a}(\bar{o}) + 
  \sum_{i \in \hood{a}}  \min_{x_i} \left[ \locCost{i}(x_i, \bar{o}_i) \wt{i}{a} + \sum_{b
      \in \summFacSet{i}{a}}
    [\locCost{i}(x_i, \vote{i}{b}) - \locCost{i}(\tilde{x}_i, \vote{i}{b})] \wt{i}{b} \right]\right\}. \label{eq.costWithoutMsg}
\end{align}

For many reasonable choices of mismatch cost functions $\locCost{i}(\cdot,
\cdot)$ and variable alphabets $\mathcal{X}_i$, the minimization over
$x_i$ requires only a small amount of summary information about the external vote and
weight vectors, $\{\vote{i}{b}\}$ and $\{\wt{i}{b}\}$.  We think of
such summary information as a message $\msg{i}{a}$ that is passed from
variable $i$ to factor $a$ and rewrite~(\ref{eq.costWithoutMsg}) as
\begin{align}
\opinVec{a}  = \argmin_{\bar{o}} 
\left\{ \selCost{a}(\bar{o}) +   \sum_{i \in \hood{a}}  
\reactiveCost{i}(\bar{o}_i, \msg{i}{a}, \wt{i}{a})
\right\}, \label{eq.costWithMsg}
\end{align}
where $\reactiveCost{i}(\bar{o}_i, \msg{i}{a}, \wt{i}{a})$ can be
interpreted as the incremental cost of factor $a$ casting its vote on
variable $i$.  It is {\em always} possible to
reexpress~(\ref{eq.costWithoutMsg}) as~(\ref{eq.costWithMsg}) simply
by setting $\msg{i}{a}$ equal to the set of $\vote{i}{b}$ and
$\wt{i}{b}$, i.e., $\msg{i}{a} = \{\vote{i}{b}, \wt{i}{b}\}$ for all
$b \in \summFacSet{i}{a}$. This transformation is thus of real
interest only when the dimension (or cardinality) of the message is
small {\em and} it is easy to compute the incremental cost
$\phi(\cdot, \cdot, \cdot)$.  In Section~\ref{sec.varTypes} we present
a number of examples of useful mismatch cost functions and variable
alphabets that meet these criteria.

\subsection{Parallel updates and simultaneous voting}
\label{subsec.parallel}
The transformation of (\ref{alg.costComp}) to (\ref{eq.costWithMsg})
enables us to perform the opinion computation relatively efficiently,
but it still accounts for an overwhelming fraction of the total
running time of PMP, particularly when the set of candidate opinions
over which the minimization runs is large. Luckily, this computation is
entirely independent for each factor and does not affect any state
external to that factor, and so lines 3--9 of Algorithm~1 can be
executed in parallel for each factor.  (The update of $\dissSet$ must
then happen in a separate loop.) Thus PMP is naturally parallelizable
to any available number of cores.

A different form of parallelism, which actually changes the behavior
of the algorithm, is for multiple factors to change their votes
simultaneously.  In the simplest version of PMP, only the single most
confident factor changes its vote during each iteration.  For this
version, we can easily guarantee convergence because at each iteration
one memory factor changes its vote vector to make a guaranteed
non-increasing change in the overall objective.  Other memory factors
are then able to adjust to the change before computing their new
opinions.

In contrast, we could let multiple factors (for example, the most
confident 10 percent) simultaneously change their votes:
($|\changeSet| > 1$) . The principal positive effect of this
modification is a reduction in the total number of times that the
computation (\ref{alg.costComp}) is executed over the course of the
algorithm. To see this, note that in the non-simultaneous version, each
factor enters the reacting set $\reactSet$ every time that a nearby
factor votes on one of its neighboring variables. With simultaneous voting,
if two nearby factors change their vote simultaneously, then our factor will
only react once in total rather than twice. Since every factor in the
reacting set participates in opinion updates, reducing the number of
times a factor enters the reacting set reduces the total number of
opinion updates, and the algorithm consumes less CPU time, even on a
single core computer. Simultaneous voting also significantly reduces the overall number
of iterations required for PMP to converge, as we shall see later.

In terms of the quality of the optimum found,
it is not clear whether simultaneous voting should be superior or inferior
to serial voting. Simultaneous voting has the potential advantage that it can make
PMP less sensitive to the whims of a single
factor. If the most confident factor disagrees with the
next four most confident factors, then letting it vote by itself is
likely to move the system in a non-optimal direction; the other
factors may now change their opinions based on the first factor's
votes, and may not get a chance to express their majority
opinion. On the other hand, if all five factors vote simultaneously, 
the system will move in a way consistent with the majority. Empirically,
we found that PMP converged to slightly better optima on the MNIST
handwritten digit classification task when simultaneous voting was used (see
Section~\ref{ssec.mnist}).  
On the other hand, simultaneous voting slightly degraded the empirical
performance of PMP on the restoration of corrupted images (see
Section~\ref{ssec.cifar}).

Simultaneous voting does introduce a complication to PMP in that it is no
longer true that the cost must be monotonically non-increasing each
iteration.  Updating factors may share variables and the change that
one factor makes in its vote vector could alter the opinion of another
factor. In practice, however, we observe that when using this
procedure the cost function increases only extremely rarely.  Further,
it is possible to recover a guarantee of convergence through the
following simple modification. We can observe whether the changes in
vote vectors in any particular iteration result in a cost increase due
to the interference between vote-changing factors.  If a cost increase
is observed, we roll back those vote changes and, for that iteration,
return to the the serial procedure, only allowing the single most
confident factor to change its votes. In practice this rollback occurs
rarely; in the MNIST task, fewer than 1\% of simultaneous votes were
retracted.

\section{Types of variables and objective functions}
\label{sec.varTypes}

In this section we consider several options for the alphabet $\mathcal{X}_i$ 
and cost $\locCost{i}$ of variable $x_i$, and particularly consider 
different cases of the minimization
\begin{equation}\label{eq.variableMin}
\optVar{i} = \argmin_{x_i} \sum_{b \in \mathcal{B}} \locCost{i}(x_i,\vote{i}{b})\wt{i}{b}
\end{equation}
with a fixed set of votes $\vote{i}{\mathcal{B}}$ for some
$\mathcal{B} \subseteq \hood{i}$.  A minimization of the above form is
used in PMP both in computing the final variable settings to return
and in computing the opinion of a factor via~\eqref{eq.costWithoutMsg}, which
is the version of~\eqref{alg.costComp} that can be given a message-passing 
interpretation.
It is thus of great importance that the minimization
in~\eqref{eq.variableMin} is tractable. Indeed for our choices of
alphabets and cost functions, we obtain explicit solutions which allows us to
write~\eqref{alg.costComp} in a simple message-passing form, per our
discussion in Sec.~\ref{sec.PMPasMsgPass}.  The messages come from all
neighboring variable nodes and summarize the votes (and associated
weights) of the memory factors neighboring each of those variables.

\subsection{Real variables with quadratic cost}

First we consider the case where $x_i \in
\mathbb{R}$ and the local cost functions are quadratic, thus
$\psi(x,v) = (x-v)^2$. Under a quadratic local cost
function, \eqref{eq.variableMin} becomes
\begin{equation*}
\optVar{i} = \argmin_{x_i} \sum_{b \in \mathcal{B}} \wt{i}{b} (x_i - \vote{i}{b})^2.
\end{equation*}
For this cost functional we take the derivative with respect to $x_i$
and set the result equal to zero to find the minimizing value of $x_i$:
\begin{equation}\label{eq.variableMin_real}
\optVar{i} = \frac{\sum_{b \in \mathcal{B}} \wt{i}{b} \vote{i}{b}}{ \sum_{b \in \mathcal{B}} \wt{i}{b}}.
\end{equation}
In other words, the minimizing $x_i$ is a weighted combination of
votes. Note that this result extends to the complex case $x_i \in \mathbb{C}$ with
$\locCost{i}(x,v) = |x - v|^2$.

We show in Appendix A that if we apply this
result to update opinion vectors for factor $a$ then
\eqref{alg.costComp} becomes
\begin{equation}\label{eq.opinionFromMessagesReal}
\opinVec{a} = \argmin_{\bar{o}} \left[ \selCost{a}(\bar{o}) + \sum_{i
    \in \hood{a}}
  \frac{\wt{i}{a}(\wtSum{i}-\wt{i}{a})}{\wtSum{i}}\left(\bar{o}_i -
  \tilde{x}_i \right)^2\right],
\end{equation}
where
\begin{equation*}
\tilde{x}_i = \frac{\sum_{b \in \summFacSet{i}{a}}
  \vote{i}{b}\wt{i}{b}}{\sum_{b \in \summFacSet{i}{a}} \wt{i}{b}}
\end{equation*}
is the weighted average of all the votes except for the one from
factor $a$, and
\begin{equation*}
\wtSum{i} = \sum_{b \in \hood{i} \setminus \abstSet}\wt{i}{b}
\end{equation*} 
is the sum of weights of non-abstaining factors. Note that $\wtSum{i}
- \wt{i}{a} = \sum_{b \in \summFacSet{i}{a}} \wt{i}{b}$.

Equation~(\ref{eq.opinionFromMessagesReal}) tells us that, in this
setting, the message is $\msg{i}{a} = \{\tilde{x}_i, \wtSum{i} -
\wt{i}{a}\}$ which tells us the value $\tilde{x}_i$ that the rest of
the network prefers, and the strength $\wtSum{i} - \wt{i}{a}$ of that
preference.

\subsection{Integer variables with linear cost}
\label{subsec.intVars}

We now consider the situation where $x_i \in \mathbb{Z}$ and the
mismatch cost $\psi : \mathbb{Z} \times \mathbb{Z} \rightarrow
\mathbb{Z}$ is absolute difference, i.e., $\psi(x,v) = |x-v|$.
Further, we consider the setting where weights are all identical. We
will argue that taking $x_i$ to be a median value of the votes
minimizes the cost. The median is actually a set of values: for example,
if there are four votes with values $1$, $2$, $3$, and $6$, the median
would be the set $\{2, 3\}$. More generally, 
the median is the set of integers for which at
least half of the votes have values no greater than the lower end of
the range and at least half the votes have values at least as large as
the upper end of the range.  Thus,
\begin{equation*}
\optVar{i} \in \mathcal{M}_{\rm med} \left(\vote{i}{\mathcal{B}} \right) = \{ v : |\{ \vote{i}{b} : \vote{i}{b} \leq v\}| \geq \lceil n/2\rceil,  |\{ \vote{i}{b} : \vote{i}{b} \geq v\}| \geq \lceil n/2\rceil, b \in \mathcal{B}\}.
\end{equation*}
We note that the median set can be summarized by its smallest and
largest elements $l$ and $u$ as $\mathcal{M}_{\rm med} = \{l, \ldots,
u\}$.  Of course, it is possible that that the set contains a single
element, in which case $l = u$.


We now argue why a minimizing $\tilde{x}_i$ must be an element of the
median set $\mathcal{M}_{\rm med}$.  First we show that
all elements in the median set have equal cost.
The set can in fact only have multiple elements if $i$ is an even-degree
variable node.  In that case half of the elements will be at most
equal to the least element of the set and half will be at least equal
to the largest element in the set.  (This is where the assumption of
identical weights is used.)  If we change the value of
$\tilde{x}_i$ among elements of the median set, the mismatch cost will
increase for half the votes the same amount it decreases for the other
half.  Thus, the votes are balanced in the range of the median set, so
we confirm that all elements in the median set have equal cost.
Next, to understand why a choice of $\tilde{x}_i$ outside of the
median set incurs greater cost, consider the case when $\tilde{x}_i$
is set equal to the largest value in the median set.  If we were to
increase $\tilde{x}_i$ further by $\Delta > 0$ the mismatch cost
$|\tilde{x}_i - \vote{i}{b}|$ would increase by at least $\Delta
(\lceil n/2 \rceil + 1)$ due to the fact that at least $\lceil
n/2\rceil$ elements have values at most equal to the lowest value in
the median set and at least one element has value equal to the largest
value in the median set.  The mismatch cost for elements with values
larger than the largest value in the median set could decrease, but at
most by $\Delta (\lceil n/2 \rceil - 1)$ and so the cost increases if
we take $\tilde{x}_i$ to be anything larger than the largest element
of the median set.  Similar logic holds if we choose $\tilde{x}_i$ to
be smaller than the smallest element in the median set.

With this solution to \eqref{eq.variableMin}, the message-passing from
of the opinion update becomes
\begin{equation}\label{eq.opinionFromMessagesMedian}
\opinVec{a} = \argmin_{\bar{o}} \left[ \selCost{a}(\bar{o}) + \sum_{i
    \in \hood{a}} \reactiveCost{i}(\bar{o}_i, \msg{i}{a}, \wt{i}{a})
  \right] = \argmin_{\bar{o}} \left[ \selCost{a}(\bar{o}) + \sum_{i
    \in \hood{a}} \wt{i}{a}d(\bar{o}_i, l_i, u_i)\right].
\end{equation}
In~(\ref{eq.opinionFromMessagesMedian}), $l_i$ and $u_i$ are the
smallest and largest element of $\mathcal{M}_{\rm
  med}\left(\vote{i}{\mathcal{B}}\right)$ where $\mathcal{B} =
\summFacSet{i}{a}$, i.e., the set of
non-abstaining memory factor neighboring variable $i$, exclusive of
$a$.  Futher, the function $d: \mathbb{Z} \times \mathbb{Z} \times
\mathbb{Z} \rightarrow \mathbb{Z}$ defined as
\begin{equation*}
d(z,l,u) = \begin{cases}
        z - u & z > u \\
        0     & l \le z \le u \\
        l - z & z < l
\end{cases}
\end{equation*} 
is the distance from $z$ to the interval $[l,u]$. 

In this case the summary message variable $i$ passes to memory factor
$a$ is the pair of integers $\msg{i}{a} = (u_i,l_i)$.  This message is
sufficient for factor $a$ to determine the impact of choosing a
particular opinion for variable $i$ on the global cost function.  The
impact is summed across all variables that memory factor $a$ neighbors.

Finally, we note that the weights $\wt{i}{a}$ appearing in Equation
(\ref{eq.opinionFromMessagesMedian}) are subject to the assumption
that all weights for the same variable are the same; e.g. $\wt{i}{a} =
\wt{i}{b}$ for all $b \in \hood{i}$.  Different weights may be
associated with different variables.

\subsection{Labels with histograms}
\label{subsec.labelVars}

We now consider a situation relevant to hypothesis testing.  In this
setting the valid $x_i$ values correspond to labels.  Generally we
consider the set of labels to be finite: $|\mathcal{X}_i| < \infty$.
As with integers we enforce the restriction that weights that are
connected to the same variable must be equal.  In this setting we take
the local cost functions to be indicators $\psi(x,v) = \mathbb{I}(x
\neq v)$.  

We minimize the cost~(\ref{eq.variableMin}) by picking $\tilde{x}_i$
to be one of the vote values in $\vote{i}{\mathcal{B}}$ that occurs
most frequently.  If we were to pick any other (less frequently
observed) vote value then the cost would increase by the decrease in
the number of votes.  Thus, it is optimal to pick $\tilde{x}_i$ to be
an element of the mode set.  which we denote as $\mathcal{M}_{\rm
  mod}(\vote{i}{\mathcal{B}})$.  As with the median set, this set may
be a singleton, or it may have multiple distinct elements.  Thus we
write that $\optVar{i}$ is optimal if
\begin{equation*}
\optVar{i} \in \mathcal{M}_{\rm mod} \left( \vote{i}{\mathcal{B}}
\right).
\end{equation*}

The message-passing form of the opinion update becomes
\begin{equation*}
\opinVec{a} = \argmin_{\bar{o}} \left[ \selCost{a}(\bar{o}) + \sum_{i
    \in \hood{a}} \wt{i}{a}\mathbb{I}\left(\bar{o}_i \not\in
  \mathcal{M}_{\rm mod}\left(\vote{i}{\mathcal{B}} \right) \right)\right],
\end{equation*}
where, as before, $\mathcal{B} = \summFacSet{i}{a}$.  In this setting
the summary message $\msg{i}{a}$ can be a binary vector of length
$|\mathcal{X}_i|$ where the non-zero coordinates indicate possible
vote values that are elements of the mode set.

\subsection{Mixing variable types}
If a factor $a$ neighbors variables of different types (real, integer,
label), the opinion update problem \eqref{alg.costComp} becomes a
combination of problem types described above.  In this situation, the
sum over $i \in \hood{a}$ can be split into sums over different types
of variables. For example, if $\hood{a} = \mathcal{B} \cup
\mathcal{C}$ with $i \in \mathcal{B}$ corresponding to complex
variables and $i \in \mathcal{C}$ corresponding to integer variables,
we can re-express~\eqref{alg.costComp} as
\begin{equation*}
\opinVec{a} = \argmin_{\bar{o}} \left[ \selCost{a}(\bar{o}) + \sum_{i
    \in \mathcal{B}}
  \frac{\wt{i}{a}(\wtSum{i}-\wt{i}{a})}{\wtSum{i}}\left|\bar{o}_i -
  \bar{x}_i \right|^2 + \sum_{i \in \mathcal{C}}\wt{i}{a}d(\bar{o}_i,
  l_i, u_i)\right].
\end{equation*}
The appropriate summary messages and update rules to use for each
variable node follow from the discussion above.

\section{Types of Memory Factors}
\label{sec.memFacTypes}

In this section we describe two ways to construct memory factors---a simple
approach using ``memory tables,'' and a somewhat more complicated, but perhaps
more scalable,
approach using reduction to a lower dimensional subspace.

\subsection{Memory Table Factors}
\label{sec.MT}
A memory table is simply a database of exemplars, each exemplar
encoding a valid configuration of the variables that neighbor the
factor in question. Memory factors that are memory tables are
trivial to train.  If we are given a number of exemplars from which to
learn local structure we simply store $x^a = \{x_i : i \in \hood{a}\}$
for each exemplar.  We think of each of these stored local snapshots as a
``memory.''  

Formally, the memory table corresponding to factor $a$ is a database
of $L_a$ (generally) distinct memories $\{\mem{a}{l} : l \in [L_a]\}$.
Each memory $\mem{a}{l}$ is a vector of length $|\hood{a}|$, the $j$th
element of which is $\memInd{a}{l}{j}$.  Each element of this vector
corresponds to a particular variable that neighbors factor $a$.  If
$\memInd{a}{l}{j}$ refers to variable $x_k$ ($k \in \varSet$ is the $j$th element
of $\hood{a}$) , then $\memInd{a}{l}{j} \in \mathcal{X}_k$.  The
memory table corresponding to factor $a$ is perhaps most conveniently
thought of as an $L_a \times |\hood{a}|$ array of variable values.

Note that evidence factors can be thought of as a special case of
memory table factors in which there is a single dynamic memory
($L_a = 1$) corresponding to the current state of the evidence.

The selection cost $\selCost{a}(\cdot)$ we use in PMP for memory table
$a$ restricts the factor's opinions (and thus votes) to memories that
are in the table.  This is accomplished by using selection cost
\begin{equation*}
\selCost{a}(\bar{o}) = \left\{ \begin{array}{ccc} 0 & \mbox{if} & \exists \ l \in [L_a] \ \mbox{s.t.} \ \bar{o} = \mem{a}{l}\\ \infty & \mbox{otherwise.} \end{array} \right.
\end{equation*}
This choice simplifies~(\ref{alg.costComp}) in the PMP algorithm.  The
outside $\arg \min$ over $\bar{o}$ restricts the optimization to
opinions that exist in memory table $a$, for which the selection cost
is zero.

The confidence $\kappa(a)$ of a memory table factor $a$ is the
difference in total cost (as specified by~(\ref{alg.costComp}))
between the best and second-best distinct sets of opinions
corresponding to its memories. Intuitively, a memory table factor has
high confidence that its choice of memory is correct, even if that
memory is costly, if the next best alternative is much
worse. Therefore it makes sense for this factor to cast its votes now
and allow other less confident factors to react to its decision.

Algorithm~1 requires us to specify the relationship
$\opinVec{a} \simeq \voteVec{a}$ denoting whether a factor's opinion
vector is sufficiently close to its vote vector for it to not want to
change those votes. Since the set of possible vote vectors in a memory
table factor is finite, we simply use equality for this comparison:
$(\opinVec{a} \simeq \voteVec{a}) \equiv (\opinVec{a} = \voteVec{a})$.

\subsection{Subspace Factors}
\label{sec.autoencoders}

Another particular form of learned structure we can utilize for memory
factors is reduction to a lower dimensional subspace (or a subset
thereof). We refer to memory factors that enforce a dimensionality
reduction as ``subspace factors.''  In particular we consider linear
subspaces, which take the form of transformations from hidden
variables $z \in \mathcal{Z}^p$ to visible variables $y \in
\mathcal{Y}^n$.  If $\mathcal{Z} = \mathbb{R}$ and $\mathcal{Y} =
\mathbb{R}$ then a subspace factor is represented using a matrix $W
\in \mathbb{R}^{n \times p}$ where the subspace is $\{y | y = Wz, z
\in \mathbb{R}^p\}$. If the variables are complex, we use complex
matrices and hidden variables. Furthermore, we often restrict
ourselves to operate on a subset of the subspace so that $\mathcal{Z}
\subseteq \mathbb{R}$ and $\mathcal{Y} \subseteq \mathbb{R}$. For
example, we will later provide an example where $\mathcal{Z} =
\mathbb{R}_+$ and $W_{ij} \in \mathbb{R}_+$.  This restricts our
subspace factor to the positive (cone) $\{y | y = Wz, z \in
\mathbb{R}^p\} \cap \mathbb{R}^n_+$.  With some abuse of terminology
we continue to refer to such memory factors as ``subspace'' factors.

Subspace factors may be learned from data by a variety of methods,
e.g.~nonnegative matrix factorization in the case of the cone
mentioned above.  Because we do not apply any sparsity assumptions to
$W$, nor to the hidden variables $z$, we generally enforce $p < n$ so
that the subspace factor indeed provides a low-dimensional
representation of the visible variables, which are constrained to lie
in some low dimensional subspace of $\mathcal{Y}^n$ spanned by the
columns of $W$.

In the context of an MFN, let us suppose a particular memory factor $a$ is a
subspace factor, which maps $p_a$ hidden variables $z \in \mathcal{Z}_a^{p_a}$
to $n_a$ visible variables $y \in \mathcal{Y}_a^{n_a}$.  We require the set of
neighboring variables $\hood{a}$ to include all visible variables represented
by the subspace factor, as the variables represented in the MFN are presumed
to be those of interest in the problem. In this paper we assume $\hood{a}$ is
precisely the set of visible variables for the subspace factor $a$.   Thus
$n_a = |\hood{a}|$ and $\mathcal{Y}_a = \mathcal{X}_i$ for $i   \in \hood{a}$.
Note that this restricts all variables in $\hood{a}$   to have the same
alphabet, a restriction that could be lifted by   allowing subspace factors to
have visible variables in different   alphabets. Such subspace factors are
possible, but we do not consider them here.

For the selection cost
$\selCost{a}$ for votes from a subspace factor, we use
\begin{equation}\label{eq.AE-selcost}
\selCost{a}(\voteVec{a}) = 
\begin{cases} 0 & \mbox{if there is a }  z \in \mathcal{Z}_a^{p_a} \st \voteVec{a} = W_az \\
            \infty & \text{ otherwise.}
\end{cases}
\end{equation}
This cost simply requires the vote $\voteVec{a}$ to be in the subspace
(or subset thereof) defined by the factor, and allows the hidden
variables to take on any value in their domain. Because the selection
cost is either zero or infinite, we can replace it in the opinion
update optimization with a feasibility constraint, so the problem
becomes
\begin{equation*}
  \opinVec{a}  = \argmin_{\bar{o},z \st \bar{o} = W_a z, \ z \in \mathcal{Z}_a^{p_a}} \left\{ \min_{x^a}
  \sum_{i \in \hood{a}} \left[ \locCost{i}(x_i, \bar{o}_i) \wt{i}{a} + 
    \sum_{b \in \summFacSet{i}{a}} \locCost{i}(x_i, \vote{i}{b}) \wt{i}{b}
  \right]\right\}.
\end{equation*}
While this is a reasonable constraint for a variety of variable types,
our examples make use of the case where $\bar{o}$ and $z$ are both
real (or both complex). Using the notation of Section
\ref{sec.varTypes}, for real variables with quadratic cost the problem
is
\begin{align}\label{eq.AE-local-prob}
\argmin \quad \sum_{i \in \hood{a}}&\frac{\wt{i}{a}(\wtSum{i}-\wt{i}{a})}{\wtSum{i}}\left(\bar{o}_i - \tilde{x}_i \right)^2 \\
\subjectto \quad\quad \bar{o} &= W_az. \notag
\end{align}
This convex quadratic program is small and straightforward to solve. Note that linear
restrictions can be included in the alphabet $\mathcal{X}_i$ without
altering the nature of this problem. One example when we restrict
variables to be non-negative, i.e., $\mathcal{X}_i= \mathbb{R}_+$.  In
this case (for which we will subsequently provide 
examples) all entries of $W_a$ are constrained to be non-negative.  In
this case the problem becomes
\begin{align*}\label{eq.AE-nonneg-local-prob}
\argmin \quad \sum_{i \in \hood{a}}&\frac{\wt{i}{a}(\wtSum{i}-\wt{i}{a})}{\wtSum{i}}\left(\bar{o}_i - \bar{x}_i \right)^2 \\
\subjectto \quad\quad \bar{o} &= W_az \\
\bar{o}, z &\ge 0.
\end{align*}
We can also extend the problem to complex variables with cost $\locCost{i}(x,v) = |x - v|^2$.

We will define the confidence $\kappa(a)$ of a subspace factor $a$ to be 
\begin{equation}\label{eq.AE-conf}
\kappa(a) = \frac{1}{|\hood{a}|}\left(\sum_{i \in \hood{a}} - \lambda \left|\summFacSet{i}{a} \right|\left|\bar{o}_{i} - \bar{x}_i\right|^2 - \frac{1}{|\hood{i} \setminus \abstSet|}\right)
\end{equation}
where $\lambda$ is a system parameter we can choose to balance the term
representing mismatches between an opinion and the incoming messages and the
term representing the number of votes those messages represent. Intuitively, a
factor should be confident if it has a lot of information from the rest of the
system and its opinion closely matches that information. We scale the score by
the total number of adjacent variables so the penalty for variables with few
votes does not have more relative impact at subspace factors with more
variables in total.

Finally, we choose some parameter $\alpha > 0$ and define the relationship
``$\simeq$'' as
\[\opinVec{a} \simeq \voteVec{a} \iff \left\lVert\opinVec{a} - \voteVec{a}\right\rVert_2^2 \le \alpha\]
where $\lVert \cdot\rVert_2$ is the $\ell^2$ norm. This means that
subspace factors will be satisfied if there is only a small difference
between their opinion vector and their vote vector. This prevents PMP
from making an infinite number of decreasingly small updates.

\section{Applications}
\label{sec.applications}

In this section we describe a number of applications that illustrate
the basic mechanisms and abilities of memory factor networks. We also compare
the behavior of MFNs using memory tables and those using subspace factors.

\subsection{Face reconstruction}
\label{ssec.faces}

Our first inference application is the reconstruction of missing or noisy data
in a two-dimensional color image. We use the FEI \cite{fei} dataset of 400 $52
\times 72$ pixel images, each showing a single face manually aligned so facial
features are in the same pixel locations on each image. We use 320 of these
images as a training set and hold out 80 for testing. We will describe a
memory factor network (with both a memory table version and a subspace
factor version) that represents such images and can be used for a
variety of tasks with similar images. The set of variables we are interested
in are the red, green, and blue pixel values for each pixel position in an
image, with an additional set of variables representing the gray value of each
pixel being added for some examples. Variables are therefore nonnegative real
numbers. They are also normalized to be at most one, but we do not enforce
this explicitly in the MFN, and simply truncate to one in the rare cases the
MFN returns a larger value.

\subsubsection{Factor layout}

The factors we use for this problem cover square grids of pixels at a
particular location in the image. One basic structure is to create three factors
(representing the red, green, and blue channels) for $8 \times 8$ squares
of pixels in the image, with squares starting every 4 pixels so each factor shares a $4
\times 4$ pixel square with three neighboring factors of the same type. At corners
and edges we follow the same pattern but truncate factors as necessary. This
structure essentially creates three parallel factor networks representing the
three color channels, since none of these factors connected to $8 \times 8$ pixels
are adjacent to variables of multiple
colors. 

To link the networks together we introduce factors that include
variables of all three colors. To keep these from being overly large we have
them cover $4 \times 4$ pixel squares (so they have $3 \cdot 16 = 48$
visible variables). Beyond limiting the total size of each factor this
structure of having ``linked'' factors with $1/4$ the size of single
channel factors greatly simplifies the connections between factors.
In our implementation it is convenient to group variables into sets and have
connections between sets of variables and factors rather than directly
between variables and factors. With this choice of the size of linked
factors all of our sets can be $4 \times 4$ grids of single channel pixel
values, and each such set thus has connections to four $8 \times 8$ mono-colored factors and
one factor that links the three colors.

We also consider another layout in which $8 \times 8$ grids of
pixels are covered by a single factor encoding all three colors, with the same
overlapping structure as the single channel factors above. This greatly
increases the size of each individual factor (a typical factor is connected 
to $8 \cdot 8 \cdot 3 = 192$ variables), but results in a network with
fewer factors and more knowledge of the connections between colors.

\subsubsection{Learning memories}

In the case of memory tables, implementation of these factors is simple: each
table includes the appropriate pixel values from each face in the training
set. For subspace factors we must learn the matrix $W$ (after choosing some order
for the variables so they are represented as a vector). Since pixel values are
nonnegative, we choose to use nonnegative matrix factorization (NMF) for this
learning step. To be precise, to learn a given matrix $W$ for a subspace factor
with $n$ variables from a training set of of $m$ images, we choose some value
$p < n$, construct an $n \times m$ matrix $X$ where the $i$th column contains
the pixel values at the appropriate positions for the $i$th training image,
and use a standard NMF algorithm to compute an approximate factorization
\[X \approx WH\] 
where $W \in \mathbb{R}_+^{n \times p}$ and $H \in \mathbb{R}_+^{p \times m}$.
In our implementation we use the NMF.jl package for the
Julia programming language and its ``Alternate Least Square
using Projected Gradient Descent'' algorithm. The matrix $W$ is the desired
subspace matrix, with $H$ representing the hidden variable values that correspond
to each of the training samples. Note that this approach works for any
arbitrary collection of scalar variables, and is not dependent on having
square samples, or on the subspace matrices 
having a particular size, or on the variables having
any particular meaning. We simply need to transform whatever portion of the
image we are interested in into a vector of scalars and then stack those
vectors into a matrix which we can factorize. The only restriction comes in
our choice of $p$, which should be smaller than $n$ to avoid trivial
factorizations. Generally we will choose $p$ much smaller than $n$ so the
subspace is in some sense representing the ``key features'' of the image
segment, as learned from the training set.

\begin{figure}[t]
\centering
\begin{subfigure}{0.23\columnwidth}
    \centering
	\includegraphics[scale=1.5]{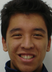}
	\caption{Original}
\end{subfigure}
\begin{subfigure}{0.23\columnwidth}
    \centering
	\includegraphics[scale=1.5]{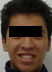}
	\caption{Evidence}
\end{subfigure}
\begin{subfigure}{0.23\columnwidth}
    \centering
	\includegraphics[scale=1.5]{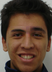}
	\caption{Memory table soln.}
\end{subfigure}
\begin{subfigure}{0.23\columnwidth}
    \centering
	\includegraphics[scale=1.5]{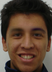}
	\caption{Subspace solution}
\end{subfigure}
\caption{An example output of the MFN. Run with $8 \times 8$  factors for each
color channel, with $4 \times 4$ linked factors, with 5 hidden variables for
all subspace factors. Factors cover a region with corners at $(7,23)$ and
$(46,40)$,  while the missing region in the evidence has corners $(9,26)$ and
$(43,38)$. Updates  were done in serial fashion, and all evidence was given
weight $20$ with each memory factor having weight $1$. The memory table solution
had mean square error 0.0111 and the subspace solution had mean square error
0.0057.}

\label{fig:example}
\end{figure}

\subsubsection{Missing data}

Our primary example problem is that of a test image in which some portion of
the image has been erased, for example the eyes (as seen in Fig.
\ref{fig:example}). In general, the algorithm is able to make reasonably
plausible reconstructions of the missing pixels, based on the pixels it does
see and its memory of other face images. Naturally, the test image was not in
the training set. In this case we provide our network with evidence from
nearby sections of the image and run the algorithm to generate pixel values
for the occluded region. Because in this case we assume the evidence that is
not erased matches the ground truth, the weight of the evidence for those
pixels  is made sufficiently large to dominate the factor votes in determining
the final variable values. Without this modification we have a tendency to
blur or distort images, particularly in the subspace case since they must be
represented by so few dimensions. Subspace networks tend to generate an
``average'' set of eyes, in which the two eyes match each other and the
surrounding face but lack any significant notable features. Memory tables,
drawing directly from a wide sample of eyes, are more likely to generate
unique-looking eyes which may or may not match each other. This can also lead
to color anomalies as there is more variation in the sorts of results each
factor votes for.

\begin{figure}[t]
\centering
\begin{subfigure}{0.23\columnwidth}
    \centering
	\includegraphics[scale=1.5]{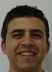}
	\caption{Original}
\end{subfigure}
\begin{subfigure}{0.23\columnwidth}
    \centering
	\includegraphics[scale=1.5]{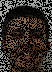}
	\caption{Evidence}
\end{subfigure}
\begin{subfigure}{0.23\columnwidth}
    \centering
	\includegraphics[scale=1.5]{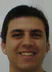}
	\caption{Memory table soln.}
\end{subfigure}
\begin{subfigure}{0.23\columnwidth}
    \centering
	\includegraphics[scale=1.5]{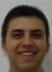}
	\caption{Subspace solution}
\end{subfigure}
\caption{An example output of an MFN with random missing evidence.  Run with
$8 \times 8$ factors for each color channel, with $4 \times 4$ linked
factors, with 5 hidden variables for all subspace factors. Factors cover the
entire image. Updates  were done using simultaneous voting, with the most confident
$50\%$ of subspace factors wanting to update doing so on  each iteration. Evidence
is given weight $2$ with memory factors given weight $1$. The memory table
solution had mean square error 0.0039 and the subspace solution had mean 
square error 0.0039.}

\label{fig:missing}
\end{figure}

One missing data instance where it may be more advantageous to
allow blurring of received data, however, is the case where data is missing from random
locations of the image rather than only in a given section. When data loss is
isolated (e.g.~the eyes) there is little to be gained from ``correcting'' the
data that is provided, but when data loss is spread among sections of the
image where most data is present, treating evidence as absolute truth can
result in anomalies as some pixels receive smoothed values while adjacent
pixels keep the more variable original. In such cases it may make more sense
to simply smooth over the whole image, which usually results in a blurry but
consistent-looking face, rather than one with odd blotches. See Fig. 
\ref{fig:missing} for an example of this blurring in an image with randomly
missing evidence.

\begin{figure}[t]
\centering
\begin{subfigure}{0.23\columnwidth}
    \centering
	\includegraphics[scale=1.5]{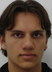}
	\caption{Original}
\end{subfigure}
\begin{subfigure}{0.23\columnwidth}
    \centering
	\includegraphics[scale=1.5]{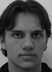}
	\caption{Evidence}
\end{subfigure}
\begin{subfigure}{0.23\columnwidth}
    \centering
	\includegraphics[scale=1.5]{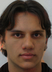}
	\caption{Memory table soln.}
\end{subfigure}
\begin{subfigure}{0.23\columnwidth}
    \centering
	\includegraphics[scale=1.5]{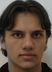}
	\caption{Subspace solution}
\end{subfigure}
\caption{An example output of the MFN. Run with $8 \times 8$ linked factors
with gray. Factors cover the entire image. Updates  were done in serial
fashion, and all (gray) evidence was given weight $100$ with each factor
having weight $1$. All subspace factors used 5 hidden variables. MFN color output
was scaled to match gray evidence. The memory table solution
had mean square error 0.0011 and the subspace solution had mean square error
0.0015.}

\label{fig:gray}
\end{figure}

Another example of missing data is the case in which the given evidence is
the grayscale pixel values rather than any color data. That is, we are given a
grayscale image of a face and would like to color it. This requires the
introduction of variables for the gray values, which will be attached to some
subset of the factors. We find that the best results come from including gray
on factors that contain all three color channels. Recall that the gray value
of a pixel can be found by taking a linear combination of its RGB values (we
use the CIE 1931 standard with $\text{Gray} = 0.212673 \cdot \text{R} + 0.715152
\cdot \text{G} + 0.072175 \cdot \text{B}$). Rather than treating this as a hard
constraint, however, the MFN treats gray values as simply another channel,
trusting to the learned structure to maintain the appropriate relationships.
For the inference problem of coloring a perfect grayscale image we add a 
post-processing step that scales the RGB pixel values to match the gray values. 
An example of the results on this problem can be seen in Fig. \ref{fig:gray}.

With colorization we see significant differences in behavior between memory
tables and subspace factors. Given strong gray evidence, clusters of memory tables
will often choose the same training face, resulting in a MFN image that
composites a relatively small number of images. The post-processing step will
then often lead to misplaced colors as, for example, pixels that were part of
the neck of a training example become part of the background of the colored
image. For subspace factors the image directly from the MFN will more closely
resemble the grayscale image, though it will be blurred. Subspace factors also
often leave sections of the colorized image gray, apparently when there is
little to distinguish between color options, whereas memory tables will always
choose a color from their memories.

\subsubsection{Noisy data}

\begin{figure}[t]
\centering
\begin{subfigure}{0.23\columnwidth}
    \centering
	\includegraphics[scale=1.5]{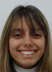}
	\caption{Original}
\end{subfigure}
\begin{subfigure}{0.23\columnwidth}
    \centering
	\includegraphics[scale=1.5]{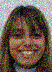}
	\caption{Evidence}
\end{subfigure}
\begin{subfigure}{0.23\columnwidth}
    \centering
	\includegraphics[scale=1.5]{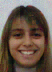}
	\caption{Memory table soln.}
\end{subfigure}
\begin{subfigure}{0.23\columnwidth}
    \centering
	\includegraphics[scale=1.5]{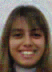}
	\caption{Subspace solution}
\end{subfigure}
\caption{An example output of a MFN with noisy evidence. Run with  $8 \times
8$ factors for each color channel, with $4 \times 4$ linked  factors, with 5
hidden variables for each subspace factor. Factors cover the entire image.
Updates  were done using simultaneous voting, with the most confident $50\%$ of factors
wanting to update doing so on  each iteration. Evidence and memory factors
were all given weight $1$. The memory table solution
had mean square error 0.0060 and the subspace solution had mean square error
0.0049.}

\label{fig:noisy}
\end{figure}

One case where it will rarely if ever make sense to privilege given evidence
is the case of noisy data. If ``known'' pixel values have been perturbed in
some way we will prefer the smoothed result given by applying our algorithm
generally, with equal weights between factors and evidence (or even extra
weight for factors). See Fig. \ref{fig:noisy} for example results in the
case of noisy data.

\subsubsection{Results summary}
We ran PMP with the described settings for each of the problems described above
on each of the 80 test images, computing the mean square error of the color-pixel
values (over the portion of the image used in the MFN). Statistics summarizing 
the results are presented in Table \ref{fig:face-summary}, and images for the
best and worst examples of each problem are available in Appendix B.

\begin{table}
\renewcommand{\arraystretch}{1.3}
\caption{Mean Square Error Result Summary for Tests of Problems Described in Fig. \ref{fig:example}-\ref{fig:noisy}}
\label{fig:face-summary}
\centering
\begin{tabular}{| c | c | c | c | c | c |}
\hline
Problem Type & Factor Type & Mean & Median & Best & Worst \\ \hline
Eyes Removed & Subspace & 0.0077 & 0.0068 & 0.0021 & 0.0210 \\ \hline
Eyes Removed & Memory Table & 0.0116 & 0.0107 & 0.0036 & 0.0346 \\ \hline
Random Dropped Evidence & Subspace & 0.0041 & 0.0039 & 0.0026 & 0.0061 \\ \hline
Random Dropped Evidence & Memory Table & 0.0055 & 0.0053 & 0.0023 & 0.0111 \\ \hline
Colorization & Subspace & 0.0020 & 0.0015 & 0.0004 & 0.0071 \\ \hline
Colorization & Memory Table & 0.0018 & 0.0016 & 0.0005 & 0.0047 \\ \hline
Noisy Evidence & Subspace & 0.0050 & 0.0049 & 0.0040 & 0.0064 \\ \hline
Noisy Evidence & Memory Table & 0.0058 & 0.0055 & 0.0035 & 0.0106 \\ \hline
\end{tabular}

\end{table}

\subsection{Music reconstruction}
\label{ssec.music}

Another application of MFNs is in the processing of audio files. Again the
question of interest may be reconstructing missing data or smoothing noisy
data, and some design decisions will be different depending on the
application, but first we will describe the general process we use for
transforming an audio file into a format appropriate for a MFN. For our data
set we use 9 second long clips of randomly generated music downloaded from
Otomata \cite{Otomata}  with 142 training
samples and 20 test samples. All clips have sample rate $40$k Hz. For memory
table factors, to limit RAM usage, a random subset of training samples was used
in each run of PMP, with each sample being included independently with probability
0.3.

\subsubsection{Creating a spectrogram}

Our MFNs will work on a spectrogram representation of the audio files, so we
use the Short-time Fourier Transform (STFT) to process our audio samples. This
process begins by splitting the audio into a series of
overlapping frames of a given length. The overlap may be varied by changing
the ``hop size'', which determines the distance between the starting time of
consecutive frames. We use a frame length of $50$ms and a hop size of
$25$ms. After splitting the frames the data in each frame is multiplied by a
window function: we use the Hanning window. Each windowed frame is then
individually Fourier transformed. After these Fourier transforms we have a matrix of complex values in
which each column represents a frame of time and each row represents a signal
frequency. The frequency range and resolution is determined by the the sample
rate of the audio and the length of the frames, respectively. In particular,
the number of frequencies represented is equal to half the number of data
points per frame plus 1, so for our $50$ms frames with the audio
sampled at $40$k Hz we will have $1001$ frequencies represented, ranging from
$0$ Hz to $20$k Hz. Rather than use the entire matrix, we will decrease it in
size by grouping frequencies into bins. For a given number of frequencies
$n_f$ and desired number of bins $n_b$ (we generally use 400) we compute a
coefficient $a$ such that
\[\sum_{j = 1}^{n_b} \lfloor e^{ja} \rfloor \approx n_f\] 
and then use $\lfloor e^{ja} \rfloor$ as the number of frequencies to include
in the $j$th bin, starting with the first bin taking the lowest $\lfloor e^{a}
\rfloor$ frequencies. For bins with multiple frequencies we simply sum the
values of all included frequencies. This logarithmic binning scheme reduces
the size of the network we work on while maintaining most of the resolution
for the lower frequencies that are of more importance to the original signal.
With this binned matrix we are prepared to build a memory factor network.

Because our variables are now complex valued, to learn subspace factors we follow
the work of Baldi and Lu \cite{baldi} and use a PCA approach. More specifically, the matrix
$W$ is formed from the $p$ eigenvectors with greatest magnitude eigenvalues of
the covariance matrix $\Sigma = \sum_{t}x_tx_t^*$ where $\{x_t\}$ are the
training vectors (note that $W$ in our notation is $A$ in the notation of
Baldi and Lu). Memory tables continue to simply store all training examples.

\subsubsection{Factor layout}

Because the perceived quality of a sound generally depends on all frequencies
present in that sound (possibly on non-adjacent frequencies), we consider memory
factors that cover the entire frequency spectrum over a small period of time.
This allows the factor to learn the entire frequency profile of a given sound,
which often includes many distant frequencies. For our main test case of
randomly generated music we also note that absolute time position has little
meaning, as we have no a priori expectation that, for example, the last two
seconds of a sample should be any different than the first two seconds.
Because of this, instead of learning different memory factors for different
parts of the spectrogram, we learn a single subspace matrix or set of memories
from all time-positions of all training examples and share this between all
memory factors in the network. Note that this results in a very large training
set even for a relatively small number of music samples.

We note that this time-independence is a feature of our particular choice of
audio signal, and would not hold true for something more structured in time,
like a short piece of speech. Such examples would more closely resemble the
face example, where specific features generally appear at the same time
position (and possibly frequency position) in every sample.

\subsubsection{Missing music}

\begin{figure}[t]
\centering
\begin{subfigure}{0.23\columnwidth}
    \centering
	\includegraphics[scale=0.27]{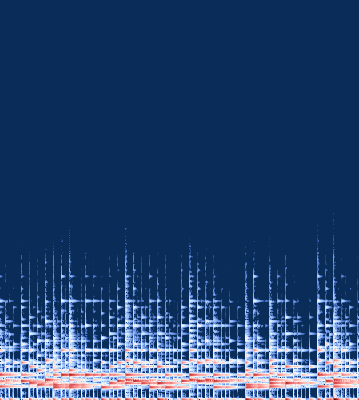}
	\caption{Original}
\end{subfigure}
\begin{subfigure}{0.23\columnwidth}
    \centering
	\includegraphics[scale=0.27]{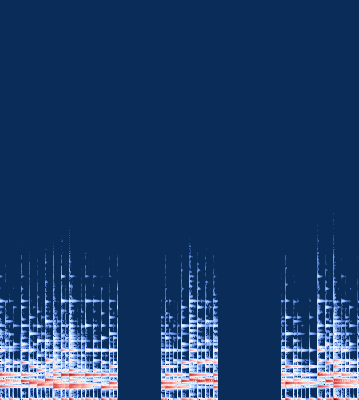}
	\caption{Evidence}
\end{subfigure}
\begin{subfigure}{0.23\columnwidth}
    \centering
	\includegraphics[scale=0.27]{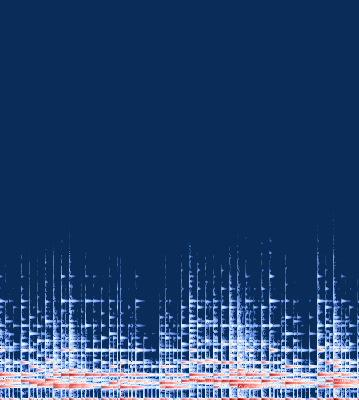}
	\caption{Memory table soln.}
\end{subfigure}
\begin{subfigure}{0.23\columnwidth}
    \centering
	\includegraphics[scale=0.27]{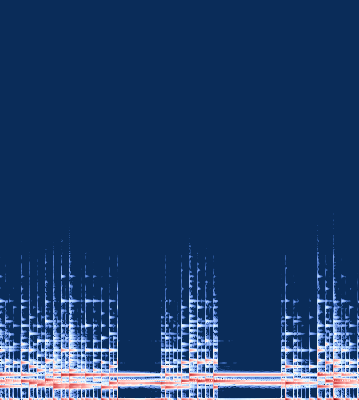}
	\caption{Subspace solution}
\end{subfigure}
\caption{An example output of an MFN on music with missing evidence.
Run with $10$ pixel wide factors covering the entire frequency range, with $5$
hidden variables for the subspace factors. Updates  were done in serial fashion.
Evidence is given weight $20$ with memory factors given weight $1$. The memory
table solution had mean square error 16.258 and the subspace solution had mean
square error 7.130.}

\label{fig:music-missing}
\end{figure}

Analogously to the removal of eyes from a face, we consider the inference
problem of filling in gaps in a piece of music. In this situation we will give
significant weight to the evidence present, preferring to maintain the original
signal where possible. It will also generally be advantageous to pick
relatively wide memory factors (10-20 spectrogram pixels) to maximize the
connection between the newly created signal and the original signal it is
adjacent to. For example spectrograms for a reconstruction problem, see Fig. 
\ref{fig:music-missing}. 

For constructing missing music, memory tables have significantly better
properties than subspace factors in our testing. The memory table network selects
full notes or sequences of notes and attempts to choose those that fit best
with the surrounding music, whereas subspace factors generally match the dominant
frequencies of the surrounding music but do not recreate the shape of notes,
rather creating a sort of flat buzz, often with a volume spike. We believe
part of the reason for this behavior is that the subspace factors are trained on a
variety of shifts in time of a note and rather than being forced to choose
between these shifts are able to combine them, creating smoother, blended
sections of spectrogram that are quite unlike the patterns of notes in the
training set.

\subsubsection{Noisy music}

Another problem we consider is that of removing noise from a music sample. To
simulate this problem we add normal random noise to our test sample. Because
many regions of the spectrogram have value zero in the absence of noise, if
evidence is given any significant weight relative to memory factors a certain
degree of noise will still be present at the end of the algorithm. To combat
this effect we run PMP twice, first with a large weight on the evidence and no
votes from factors and then again with very little weight for the evidence but
taking the final votes of the factors as the initial votes in the second run.
This allows factors to take advantage of the evidence while not allowing it to
dominate. Because factors primarily need to match the signal underneath the
noise rather than coordinating over time intervals, we use very narrow factors (2 pixels)
for noise problems.

\begin{figure}[t]
\centering
\begin{subfigure}{0.23\columnwidth}
    \centering
	\includegraphics[scale=0.27]{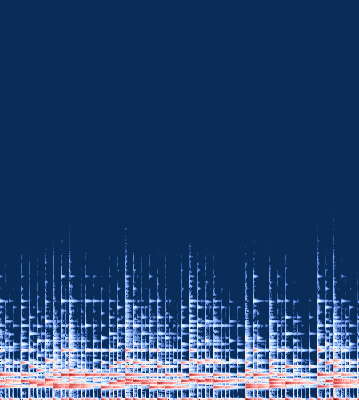}
	\caption{Original}
\end{subfigure}
\begin{subfigure}{0.23\columnwidth}
    \centering
	\includegraphics[scale=0.27]{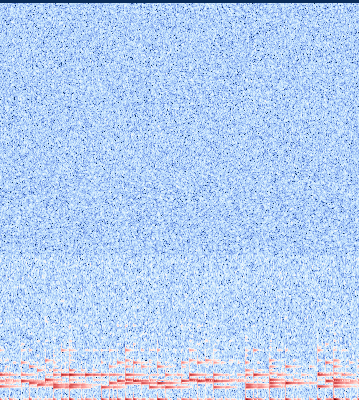}
	\caption{Evidence}
\end{subfigure}
\begin{subfigure}{0.23\columnwidth}
    \centering
	\includegraphics[scale=0.27]{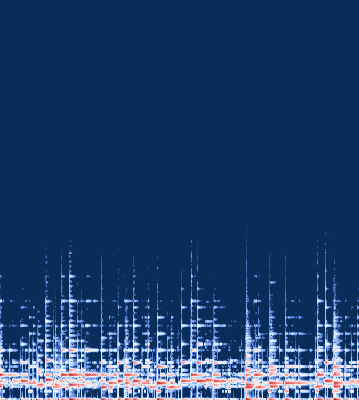}
	\caption{Memory table soln.}
\end{subfigure}
\begin{subfigure}{0.23\columnwidth}
    \centering
	\includegraphics[scale=0.27]{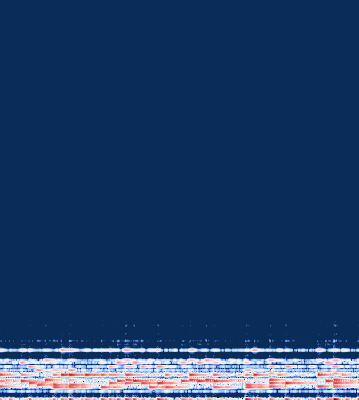}
	\caption{Subspace solution}
\end{subfigure}

\caption{An example output of an MFN on music with noisy evidence. Run with
$2$ pixel wide factors covering the entire frequency range, with 20 hidden
variables for the subspace factors. Updates  were done in serial fashion. Evidence
is originally given weight $20$ with memory factors given weight $1$, with
evidence weight set to $0.01$ for the second run initialized with the final
factor votes of the first. The memory table solution had mean square error 
10.386 and the subspace solution had mean square error 1.212.}

\label{fig:music-noisy}
\end{figure}

An example of a noisy music problem can be seen in Fig. \ref{fig:music-noisy}.
Again we see that memory tables are better able
to maintain the rich structure of notes, though subspace factors do maintain much
more of the structure of music than they did when attempting to create
stretches of it, and are better able to exactly match the main structures than
memory tables, which must try to find matching notes among their memories.
Combining these notes from different training samples results in a somewhat
more disjointed reconstruction, whereas subspace factors are somewhat fuzzier but
smoother.

\subsubsection{Results summary}
We ran PMP with the described settings for the two problems described above
on each of the 20 test clips, computing the mean square error of the spectrogram 
pixel values. Statistics summarizing the results are presented in Table
\ref{fig:music-summary}, and images for the best and worst examples of each 
problem are available in Appendix B.

\begin{table}
\renewcommand{\arraystretch}{1.3}
\caption{Mean Square Error Result Summary for Tests of Problems Described in Fig. \ref{fig:music-missing} and \ref{fig:music-noisy}}
\label{fig:music-summary}
\centering
\begin{tabular}{| c | c | c | c | c | c |}
\hline
Problem Type & Factor Type & Mean & Median & Best & Worst \\ \hline
Sections Removed & Subspace & 8.248 & 7.270 & 3.712 & 20.366 \\ \hline
Sections Removed & Memory Table & 14.025 & 12.835 & 8.205 & 28.042 \\ \hline
Noisy Evidence & Subspace & 1.459 & 1.238 & 0.635 & 4.303 \\ \hline
Noisy Evidence & Memory Table & 13.100 & 10.399 & 4.932 & 42.328 \\ \hline
\end{tabular}
\end{table}

\subsection{Handwritten digit classification}
\label{ssec.mnist}

We now demonstrate the use of label variables and mixed-variable
factors by using memory table MFNs to classify handwritten digits. We
use the MNIST dataset \cite{MNIST} consisting of 60,000 training
examples of $28\times28$ grayscale images with correct classifications
and 10,000 test images with no classification.

We work with $32\times32$ images, in which the original images have
been centered, in order to facilitate a hierarchy of pixel and label
variables. In addition to $32^2$ level-0 variables representing the
image's pixels, there are $16^2$ hidden level-1 variables
corresponding to a lower-resolution version of the same image, $8^2$
hidden level-2 variables corresponding to an even lower resolution
version, and $4^2$ hidden level-3 variables representing the entire
image. A hierarchy of label variables exists in parallel to the
hierarchy of pixel variables, with $15^2$ level-0 variables, $7^2$
level-1 variables, $3^2$ level-2 variables, and a single level-3
variable characterizing the entire image. The pixel variables are
integer variables with linear cost and weights of 1, while the label
variables use an indicator cost as described in
Section~\ref{subsec.labelVars} with weights of 32. The label variables
have a higher weight than the pixel variables both because there are
fewer of them and because the natural scale of a mismatch is lower.

This hierarchy has two useful properties. Firstly, it allows
locally inferred labels to be synthesized into one global label which
can then be read off as the classification of the entire
image. Secondly, it provides a ``fast lane'' for information to
propagate from one part of the image to a more remote part. Inferences
about pixels or labels can flow up and down the hierarchy in a faster
way than if they had to linearly diffuse through the network.

\begin{figure}
\centering
\begin{tikzpicture}
  \newcommand\SQUARE{0.1}       
  \newcommand\SIZE{32*\SQUARE}   
  \newcommand\XPAD{8*\SQUARE}   
  \newcommand\YPAD{8*\SQUARE}   


  \foreach \x/\y/\val in {
10/12/224, 10/13/176, 10/14/169, 10/15/80, 10/16/169, 10/17/237,
11/10/227, 11/11/111, 11/12/14, 11/13/1, 11/14/2, 11/15/2, 11/16/2, 11/17/17, 11/18/187,
12/9/164, 12/10/20, 12/11/2, 12/12/2, 12/13/62, 12/14/175, 12/15/100, 12/16/47, 12/17/2, 12/18/15, 12/19/187,
13/8/241, 13/9/31, 13/10/1, 13/11/1, 13/12/47, 13/16/215, 13/17/1, 13/18/0, 13/19/97,
14/8/106, 14/9/2, 14/10/2, 14/11/2, 14/12/70, 14/13/247, 14/16/170, 14/17/2, 14/18/1, 14/19/113,
15/8/62, 15/9/2, 15/10/2, 15/11/2, 15/12/2, 15/13/157, 15/16/89, 15/17/2, 15/18/110, 15/19/243,
16/8/250, 16/9/236, 16/10/89, 16/11/2, 16/12/2, 16/13/36, 16/14/73, 16/15/21, 16/16/4, 16/17/70, 16/18/247,
17/10/237, 17/11/17, 17/12/2, 17/13/1, 17/14/2, 17/15/2, 17/16/18, 17/17/219,
18/11/140, 18/12/24, 18/13/0, 18/14/1, 18/15/1, 18/16/1, 18/17/39, 18/18/121, 18/19/246,
19/11/249, 19/12/231, 19/13/54, 19/14/64, 19/15/120, 19/16/19, 19/17/2, 19/18/1, 19/19/80, 19/20/167, 19/21/253,
20/13/237, 20/14/243, 20/16/213, 20/17/103, 20/18/32, 20/19/2, 20/20/2, 20/21/89, 20/22/207,
21/18/220, 21/19/94, 21/20/29, 21/21/2, 21/22/10, 21/23/140, 21/24/234,
22/20/237, 22/21/129, 22/22/2, 22/23/1, 22/24/37, 22/25/225,
23/22/217, 23/23/83, 23/24/1, 23/25/1, 23/26/69, 23/27/241,
24/23/246, 24/24/209, 24/25/90, 24/26/11, 24/27/32,
25/26/213, 25/27/177
  } {
    \pgfmathsetmacro\k{(255 - \val) / 2.55}
    \fill[color=black!\k] (\x*\SQUARE,         \SIZE+\YPAD+\SIZE-\y*\SQUARE-\SQUARE)
                rectangle (\x*\SQUARE+\SQUARE, \SIZE+\YPAD+\SIZE-\y*\SQUARE);
  }
  \draw[thick]        (0, \SIZE+\YPAD) rectangle (\SIZE, \SIZE*2+\YPAD);

  \foreach \x/\y/\val in {
5/5/223, 5/6/116, 5/7/81, 5/8/95, 5/9/223,
6/4/190, 6/5/18, 6/6/140, 6/7/222, 6/8/57, 6/9/64,
7/4/161, 7/5/52, 7/6/245, 7/8/89, 7/9/117,
8/4/252, 8/5/76, 8/6/42, 8/7/31, 8/8/57, 8/9/249,
9/5/227, 9/6/61, 9/7/12, 9/8/47, 9/9/111, 9/10/215, 9/11/254,
10/8/207, 10/9/91, 10/10/32, 10/11/86, 10/12/233,
11/10/228, 11/11/127, 11/12/27, 11/13/187,
12/12/196, 12/13/90
  } {
    \pgfmathsetmacro\k{(255 - \val) / 2.55}
    \fill[color=black!\k] (\SIZE+\XPAD+2*\x*\SQUARE,         \SIZE+\YPAD+\SIZE-2*\y*\SQUARE-2*\SQUARE)
                rectangle (\SIZE+\XPAD+2*\x*\SQUARE+2*\SQUARE, \SIZE+\YPAD+\SIZE-2*\y*\SQUARE);
  }
  \draw[thick]        (\SIZE+\XPAD, \SIZE+\YPAD) rectangle (2*\SIZE+\XPAD, \SIZE*2+\YPAD);

  \foreach \x/\y/\val in {
2/2/247, 2/3/177, 2/4/195,
3/2/110, 3/3/219, 3/4/108,
4/2/189, 4/3/52, 4/4/119, 4/5/233,
5/3/249, 5/4/189, 5/5/123, 5/6/170,
6/6/227
  } {
    \pgfmathsetmacro\k{(255 - \val) / 2.55}
    \fill[color=black!\k] (2*\SIZE+2*\XPAD+4*\x*\SQUARE, \SIZE+\YPAD+\SIZE-4*\y*\SQUARE-4*\SQUARE)
                rectangle (2*\SIZE+2*\XPAD+4*\x*\SQUARE+4*\SQUARE, \SIZE+\YPAD+\SIZE-4*\y*\SQUARE);
  }
  \draw[thick]        (2*\SIZE+2*\XPAD, \SIZE+\YPAD) rectangle (3*\SIZE+2*\XPAD, \SIZE*2+\YPAD);

  \foreach \x/\y/\val in {
1/1/188, 1/2/203,
2/1/186, 2/2/194, 2/3/221
  } {
    \pgfmathsetmacro\k{(255 - \val) / 2.55}
    \fill[color=black!\k] (3*\XPAD+3*\SIZE+8*\x*\SQUARE, \SIZE+\YPAD+\SIZE-8*\y*\SQUARE-8*\SQUARE)
                rectangle (3*\XPAD+3*\SIZE+8*\x*\SQUARE+8*\SQUARE, \SIZE+\YPAD+\SIZE-8*\y*\SQUARE);
  }
  \draw[thick]           (3*\SIZE+3*\XPAD,\SIZE+\YPAD) rectangle (4*\SIZE+3*\XPAD, \SIZE*2+\YPAD);

  \draw[thick]                     (0,0)       rectangle (\SIZE,\SIZE);
  \foreach \x/\y/\val in {
0/0/9, 0/1/9, 0/2/9, 0/3/9, 0/4/9, 0/5/9, 0/6/9, 0/7/9, 0/8/9, 0/9/9, 0/10/9, 0/11/9, 0/12/9, 0/13/9, 0/14/9,
1/0/9, 1/1/9, 1/2/9, 1/3/9, 1/4/9, 1/5/9, 1/6/9, 1/7/9, 1/8/9, 1/9/9, 1/10/9, 1/11/9, 1/12/9, 1/13/9, 1/14/9,
2/0/9, 2/1/9, 2/2/9, 2/3/9, 2/4/9, 2/5/9, 2/6/9, 2/7/9, 2/8/9, 2/9/9, 2/10/9, 2/11/9, 2/12/9, 2/13/9, 2/14/9,
3/0/9, 3/1/9, 3/2/9, 3/3/9, 3/4/9, 3/5/9, 3/6/9, 3/7/9, 3/8/9, 3/9/9, 3/10/9, 3/11/9, 3/12/9, 3/13/9, 3/14/9,
4/0/9, 4/1/9, 4/2/9, 4/3/9, 4/4/9, 4/5/9, 4/6/9, 4/7/9, 4/8/9, 4/9/9, 4/10/9, 4/11/9, 4/12/9, 4/13/9, 4/14/9,
5/0/9, 5/1/9, 5/2/9, 5/3/9, 5/4/9, 5/5/9, 5/6/9, 5/7/9, 5/8/9, 5/9/9, 5/10/9, 5/11/9, 5/12/9, 5/13/9, 5/14/9,
6/0/2, 6/1/2, 6/2/2, 6/3/2, 6/4/9, 6/5/9, 6/6/9, 6/7/9, 6/8/9, 6/9/4, 6/10/9, 6/11/9, 6/12/9, 6/13/9, 6/14/9,
7/0/2, 7/1/2, 7/2/2, 7/3/2, 7/4/2, 7/5/9, 7/6/9, 7/7/9, 7/8/4, 7/9/4, 7/10/4, 7/11/9, 7/12/9, 7/13/9, 7/14/9,
8/0/2, 8/1/2, 8/2/2, 8/3/2, 8/4/2, 8/5/3, 8/6/9, 8/7/8, 8/8/8, 8/9/4, 8/10/9, 8/11/9, 8/12/9, 8/13/4, 8/14/4,
9/0/2, 9/1/2, 9/2/2, 9/3/2, 9/4/2, 9/5/3, 9/6/3, 9/7/8, 9/8/8, 9/9/8, 9/10/8, 9/11/9, 9/12/4, 9/13/4, 9/14/4,
10/0/2, 10/1/2, 10/2/2, 10/3/2, 10/4/3, 10/5/3, 10/6/3, 10/7/3, 10/8/8, 10/9/8, 10/10/8, 10/11/4, 10/12/4, 10/13/4, 10/14/4,
11/0/3, 11/1/3, 11/2/3, 11/3/3, 11/4/3, 11/5/3, 11/6/3, 11/7/3, 11/8/3, 11/9/8, 11/10/4, 11/11/4, 11/12/4, 11/13/4, 11/14/4,
12/0/3, 12/1/3, 12/2/3, 12/3/3, 12/4/3, 12/5/3, 12/6/3, 12/7/3, 12/8/3, 12/9/4, 12/10/4, 12/11/4, 12/12/4, 12/13/4, 12/14/4,
13/0/3, 13/1/3, 13/2/3, 13/3/3, 13/4/3, 13/5/3, 13/6/3, 13/7/3, 13/8/3, 13/9/4, 13/10/4, 13/11/4, 13/12/4, 13/13/4, 13/14/4,
14/0/3, 14/1/3, 14/2/3, 14/3/3, 14/4/3, 14/5/3, 14/6/3, 14/7/3, 14/8/3, 14/9/4, 14/10/4, 14/11/4, 14/12/4, 14/13/4,
14/14/4
  } {
    \draw[font=\tiny] (2*\x*\SQUARE+2*\SQUARE, \SIZE-2*\SQUARE-2*\y*\SQUARE) node{\val};
  }

  \draw[thick]                     (\SIZE+\XPAD,0)       rectangle (2*\SIZE+\XPAD,\SIZE);
  \foreach \x/\y/\val in {
0/0/9, 0/1/9, 0/2/9, 0/3/9, 0/4/9, 0/5/9, 0/6/9,
1/0/9, 1/1/9, 1/2/9, 1/3/9, 1/4/9, 1/5/9, 1/6/9,
2/0/9, 2/1/9, 2/2/9, 2/3/9, 2/4/9, 2/5/9, 2/6/9,
3/0/2, 3/1/2, 3/2/9, 3/3/9, 3/4/9, 3/5/9, 3/6/9,
4/0/2, 4/1/2, 4/2/9, 4/3/9, 4/4/9, 4/5/9, 4/6/4,
5/0/3, 5/1/3, 5/2/3, 5/3/3, 5/4/4, 5/5/4, 5/6/4,
6/0/3, 6/1/3, 6/2/3, 6/3/3, 6/4/4, 6/5/4, 6/6/4
  } {
    \draw (\SIZE+\XPAD+4*\x*\SQUARE+4*\SQUARE, \SIZE-4*\SQUARE-4*\y*\SQUARE) node{\val};
  }

  \draw[thick]                     (2*\SIZE+2*\XPAD,0)       rectangle (3*\SIZE+2*\XPAD,\SIZE);

  \foreach \x/\y/\val in {
0/0/9, 0/1/9, 0/2/9,
1/0/9, 1/1/9, 1/2/9,
2/0/3, 2/1/9, 2/2/4
  } {
    \draw[font=\huge] (2*\SIZE+2*\XPAD+8*\x*\SQUARE+8*\SQUARE, \SIZE-8*\SQUARE-8*\y*\SQUARE) node{\val};
  }

  \draw[thick]   (3*\SIZE+3*\XPAD,0) rectangle (4*\SIZE+3*\XPAD,\SIZE);
  \draw[font=\Huge] (3.5*\SIZE+3*\XPAD,.5*\SIZE) node[scale=3]{9};

  \newcommand\MEMX{4}
  \newcommand\MEMY{0}
  \draw[thick, dashed] (\SIZE+\XPAD+2*\MEMX*\SQUARE,
                \SIZE+\YPAD+\SIZE-2*\MEMY*\SQUARE)
               rectangle (\SIZE+\XPAD+2*\MEMX*\SQUARE+16*\SQUARE,
                          \SIZE+\YPAD+\SIZE-2*\MEMY*\SQUARE-16*\SQUARE);
  \draw[thick, dashed] (2*\SIZE+2*\XPAD+2*\MEMX*\SQUARE,
                \SIZE+\YPAD+\SIZE-2*\MEMY*\SQUARE)
               rectangle (2*\SIZE+2*\XPAD+2*\MEMX*\SQUARE+16*\SQUARE,
                          \SIZE+\YPAD+\SIZE-2*\MEMY*\SQUARE-16*\SQUARE);
  \draw[thick, dashed] (\SIZE+\XPAD+2*\MEMX*\SQUARE+2*\SQUARE,
                \SIZE-2*\MEMY*\SQUARE-2*\SQUARE)
               rectangle (\SIZE+\XPAD+2*\MEMX*\SQUARE+14*\SQUARE,
                          \SIZE-2*\MEMY*\SQUARE-14*\SQUARE);
  \draw[thick, dashed] (2*\SIZE+2*\XPAD+2*\MEMX*\SQUARE+4*\SQUARE,
                \SIZE-2*\MEMY*\SQUARE-4*\SQUARE)
               rectangle (2*\SIZE+2*\XPAD+2*\MEMX*\SQUARE+12*\SQUARE,
                          \SIZE-2*\MEMY*\SQUARE-12*\SQUARE);

  \newcommand\RAD{.4cm*\XPAD}
  \coordinate (MT) at (2*\SIZE+1.5*\XPAD, \SIZE+.5*\YPAD);
  \draw[thick, dashed, fill=black!25] (MT) circle [radius=.4*\XPAD];

  \coordinate (P1) at (\SIZE+\XPAD+2*\MEMX*\SQUARE+16*\SQUARE,
                       \SIZE+\YPAD+\SIZE-2*\MEMY*\SQUARE-16*\SQUARE);
  \coordinate (P2) at (2*\SIZE+2*\XPAD+2*\MEMX*\SQUARE,
                       \SIZE+\YPAD+\SIZE-2*\MEMY*\SQUARE-16*\SQUARE);
  \coordinate (P3) at (\SIZE+\XPAD+2*\MEMX*\SQUARE+14*\SQUARE,
                       \SIZE-2*\MEMY*\SQUARE-2*\SQUARE);
  \coordinate (P4) at (2*\SIZE+2*\XPAD+2*\MEMX*\SQUARE+4*\SQUARE,
                       \SIZE-2*\MEMY*\SQUARE-4*\SQUARE);

  \draw[thick, dashed] (P1) -- ($(MT)!\RAD!(P1)$);
  \draw[thick, dashed] (P2) -- ($(MT)!\RAD!(P2)$);
  \draw[thick, dashed] (P3) -- ($(MT)!\RAD!(P3)$);
  \draw[thick, dashed] (P4) -- ($(MT)!\RAD!(P4)$);

  \draw (0.5*\SIZE, 2*\SIZE + 2*\YPAD) node{Level 0};
  \draw (1.5*\SIZE+\XPAD, 2*\SIZE + 2*\YPAD) node{Level 1};
  \draw (2.5*\SIZE+2*\XPAD, 2*\SIZE + 2*\YPAD) node{Level 2};
  \draw (3.5*\SIZE+3*\XPAD, 2*\SIZE + 2*\YPAD) node{Level 3};
\end{tikzpicture}

\caption{All 90 variables connected to a sample memory table factor at
  level 1. 64 variables are a high-resolution representation of an
  $8 \times 8$ region, 16 are a lower-resolution representation of the
  same region, 9 are labels correlated with $4 \times 4$ subregions of
  the high-resolution image, and one is a label correlated with the
  entire region of the low-resolution image.}
\label{fig:MNIST-factor}
\end{figure}
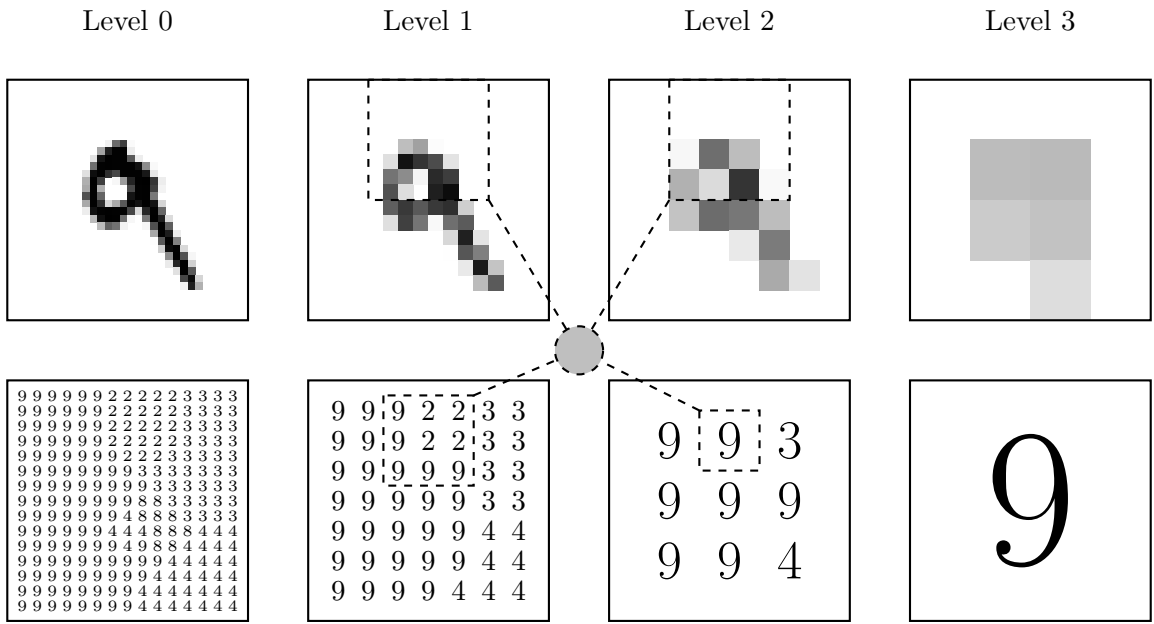

The network hierarchy is constructed in the following manner. Each
factor at level $n$ is connected to $8^2$ pixel variables and $3^2$
label variables of level $n$, as well as to $4^2$ pixel variables and
1 label variable of level $n+1$. A sample set of variables connected
to a single memory table factor is shown in
Fig.~\ref{fig:MNIST-factor}.

When a training example is presented
to the system, all label variables are set to the correct label and
all hidden pixel variables are set to the average of the pixel values
in the corresponding $2\times2$ patch of the image at the next lower
level. Every memory in each memory table thus is filled with specified
values. 

At testing time, the hidden pixel and label variables are not
connected to evidence factors; they only acquire values as a result of
being voted on by memory table factors. A level-0 factor that thinks
that a group of pixels is characteristic of a 4 will cast votes for
its neighboring level-1 pixel variables with a blurred version of its
memory and cast a vote of 4 for its neighboring level-1 label
variable. Factors pay a mismatch cost for disagreeing both with pixel
variables and with label variables, and thus a factor that isn't sure
whether it is looking at a 4 or a 9 may be swayed by the fact that a
neighboring factor has cast a vote for the label 4. When the algorithm
has converged, we classify the image according to the single label at
the top of the hierarchy.

This simple Memory Factor Network is sufficient to classify the 10,000
test images of the MNIST data set with 96.15\% accuracy when run with
a single factor voting at a time, and 96.41\% accuracy when using
simultaneous votes of the top 10\% of factors (see
Section~\ref{subsec.parallel}). On average the single-vote
implementation required 78.7 iterations and 1148.9 opinion
updates, while the simultaneous-vote implementation required 16.9
iterations and 687.2 opinion updates.

\subsection{Restoration of corrupted images}
\label{ssec.cifar}

A natural application for MFNs using memory tables is to reconstruct
previously-seen images when presented with versions of them that have
been corrupted by noise or erasure. We use a similar network to that
of the face reconstruction application and for data we use the
CIFAR-10 dataset of $32\times32$ RGB images \cite{krizhevsky},
containing 50,000 images. The provided labeling of the images is
irrelevant to our problem.

After all 50,000 images are read, one of the previously-seen images is
selected at random and has Gaussian noise applied to every color
channel value of every pixel with a standard deviation of 40. In
addition, a randomly-generated blob of 144 pixels is completely erased
from the center of the image (providing no evidence). This image is
then presented to the MFN, which restores the image to its original
version, removing the applied noise and filling in the correct pixels
of the erased region.

This application is particularly convenient because
it has a quantitative metric for assessing the quality
of the result; we simply measure the L1 distance between the computed
image and the original. This makes it a natural choice for evaluating
the effect of changes to the algorithm.

Thus, to assess the effects of simultaneous voting, 
we ran the PMP algorithm over 1000 test images using
both the standard serial algorithm and the simultaneous voting version
described in Section~\ref{subsec.parallel}, running 10\% of all
factors every iteration. While the simultaneous voting version was
less likely to reproduce the initial image exactly, its
average error was comparable to that of the serial version.  The
serial version restored 91.9\% of the images perfectly with an average
total L1 error across all color channels of 895.1 (0.291 per pixel per
channel), while the simultaneous voting version restored only 83.1\% of the
images perfectly but with an average total L1 error of 936.7 (0.305
per pixel per channel).

\section{Conclusions and future work}
\label{sec.conclusion}

We have introduced here a new approach to combining inference with learning
from experience. Memory factor networks provide a simple way to store examples
learned from experience, while proactive message passing using a confidence-based
scheme for prioritizing factor updates gives a reliable way to converge to
good optima of the memory factor network cost function. 

We consider the
algorithms and the applications demonstrated here to be just an initial foray
into the possibilities raised by this approach. Thus, one might consider
factor graphs that combined memory factor nodes with more conventional 
factor nodes that encoded known statistical dependencies or constraints. The
PMP algorithm is well-suited to MFNs, but one might nevertheless consider
using other approaches, such as those based on belief propagation or variational
approaches \cite{Wainwright}, 
or those based on the alternating directions method of multipliers 
\cite{Boyd, TWA}. There are many potential applications
that one might attack with this approach, including just from computer
vision the problems of inferring depth from single
images, or motion from videos. Finally, 
an important open question is how well the approach
described here can take advantage of massive amounts of data, as are often now
available \cite{data}.

\appendix
\section*{Appendix A: Derivation of opinion update for real variables with quadratic cost}

\label{app.real_derivation}

In this section we derive the messages in the message-passing version
of PMP for real variables and quadratic costs.  Under these choices,
(\ref{alg.costComp}) specializes to
\begin{equation}
{\opinVec{a}  = \argmin_{\bar{o}} \left\{ \selCost{a}(\bar{o}) + 
  \sum_{i \in \hood{a}} \min_{x_i} \left[ \wt{i}{a} (x_i - \bar{o}_i)^2 + 
    \sum_{b \in \summFacSet{i}{a}}  \wt{i}{b} (x_i - \vote{i}{b})^2 
  \right]\right\}}. \label{eq.recapCost}
\end{equation}
where we recall the definition of $\summFacSet{i}{a} = (\hood{i}
\backslash\{a\}) \backslash \abstSet$, originally made
in~(\ref{eq.summFacSet}).

We first optimize the choice of $x_i$ by taking the derivative with
respect to $x_i$:
\begin{equation*}
\bar{x}_i = \argmin_x \left[ \wt{i}{b} (x - \bar{o}_i)^2 + \sum_{b \in
    \summFacSet{i}{a}} \wt{i}{b} (x - \vote{i}{b})^2 \right] = \frac{\bar{o}_i\wt{i}{a} + \sum_{b
      \in \summFacSet{i}{a}}\vote{i}{b}\wt{i}{b}}{\sum_{b \in
      \hood{i}}\wt{i}{b}}.
\end{equation*}
If we define $\wtSum{i} = \sum_{b \in \hood{i} \setminus
  \abstSet}\wt{i}{b}$ then $\sum_{b \in \summFacSet{i}{a}}\wt{i}{b} =
\wtSum{i} - \wt{i}{a}$.  Multiplying and dividing the second term by
$(\wtSum{i} - \wt{i}{a})$ we get
\begin{equation*}
\bar{x}_i = \frac{\wt{i}{a} \bar{o}_i + \left(\wtSum{i}-\wt{i}{a}\right)\tilde{x}_i}{\wtSum{i}},
\end{equation*}
where 
\begin{equation}
\tilde{x}_i = \argmin_x \sum_{b \in
    \summFacSet{i}{a}} \wt{i}{b} (x - \vote{i}{b})^2  = \frac{\sum_{b
      \in \summFacSet{i}{a}}\vote{i}{b}\wt{i}{b}}{\sum_{b \in
      \summFacSet{i}{a}}\wt{i}{b}} = \frac{\sum_{b
      \in \summFacSet{i}{a}}\vote{i}{b}\wt{i}{b}}{\wtSum{i}-\wt{i}{a}} \label{eq.defTildeX}
\end{equation}
is the minimization of $x_i$ with respect solely to memory factors in
$\summFacSet{i}{a}$.  This is the same $\tilde{x}_i$ as was defined in
Sec.~\ref{sec.PMPasMsgPass}.

We denote the cost associated with a particular variable $i \in \hood{a}$ 
for a given opinion $\bar{o}_i$ as $c(\bar{o}_i)$. Then we have
\begin{align*}
c(\bar{o}_i) &= \wt{i}{a} \left(\bar{x}_i - \bar{o}_i\right)^2 +
\sum_{b \in
  \summFacSet{i}{a}}\wt{i}{b}\left(\bar{x}_i-\vote{i}{b}\right)^2
\\ &= \wt{i}{a} \left(\frac{\wt{i}{a}\bar{o}_i +
  \left(\wtSum{i}-\wt{i}{a}\right)\tilde{x}_i}{\wtSum{i}} -
\bar{o}_i\right)^2 + \sum_{b \in
  \summFacSet{i}{a}} \wt{i}{b} \left(\frac{\wt{i}{a}\bar{o}_i +
  \left(\wtSum{i}-\wt{i}{a}\right)\tilde{x}_i}{\wtSum{i}}-\vote{i}{b}\right)^2\\
&= \frac{ \wt{i}{a} (\wtSum{i} - \wt{i}{a})^2}{\wtSum{i}^2} (\tilde{x}_i - \bar{o}_i)^2 
+ \frac{1}{\wtSum{i}^2}\bigg( \sum_{b \in
  \summFacSet{i}{a} }\wt{i}{b}\left(\wt{i}{a}\bar{o}_i +
\left(\wtSum{i} - \wt{i}{a}\right) \tilde{x}_i-
\wtSum{i}\vote{i}{b}\right)^2\bigg) \\ 
&= \frac{ \wt{i}{a} (\wtSum{i} - \wt{i}{a})^2}{\wtSum{i}^2} (\bar{o}_i^2 - 2  \bar{o}_i\tilde{x}_i)
+ \frac{1}{\wtSum{i}^2}\bigg( \sum_{b \in
  \summFacSet{i}{a} }\wt{i}{b} [(\wt{i}{a}\bar{o}_i)^2 +
\wt{i}{a} \left(\wtSum{i} - \wt{i}{a}\right) \bar{o}_i \tilde{x}_i-
\wt{i}{a} \wtSum{i}\vote{i}{b}\bar{o}_i] \bigg)  + C,
\end{align*}
where the constant term $C$ includes all the terms that are not a
function of $\bar{o}$.  We next simplify the sums over $b \in
\summFacSet{i}{a}$ by recalling that $\sum_{b \in \summFacSet{i}{a}}
\wt{i}{b} = (\wtSum{i} - \wt{i}{a})$ and that,
from~(\ref{eq.defTildeX}), $\sum_{b \in \summFacSet{i}{a}} \wt{i}{b}
\vote{i}{b}= (\wtSum{i} - \wt{i}{a}) \tilde{x}_i$.  Hence,
\begin{align*}
c(\bar{o}_i) &= \frac{ \wt{i}{a} (\wtSum{i} - \wt{i}{a})^2}{\wtSum{i}^2} (\bar{o}_i^2 - 2  \bar{o}_i\tilde{x}_i)
+ \bar{o}_i^2 \frac{(\wt{i}{a})^2 \left(\wtSum{i} - \wt{i}{a}\right)}{\wtSum{i}^2}
+
\bar{o}_i \tilde{x}_i\frac{\wt{i}{a} \left(\wtSum{i} - \wt{i}{a}\right)^2}{\wtSum{i}^2} -
\bar{o}_i \tilde{x}_i\frac{\wt{i}{a}\wtSum{i}(\wtSum{i}-\wt{i}{a})}{\wtSum{i}^2}   + C\\
&= \frac{\wt{i}{a}(\wtSum{i}-\wt{i}{a})}{\wtSum{i}}\left(\bar{o}_i^2 -
2\tilde{x}_i\bar{o}_i\right) + C,
\end{align*}
where in the second equation a number of terms cancel.  We substitute
this result into~(\ref{eq.recapCost}) noting that the constant $C$ can
be dropped and adding a constant to make $\left(\bar{o}_i^2 -
2\tilde{x}_i\bar{o}_i\right)$ a quadratic form.  This yields the final
form of our optimization,
\begin{equation}
{\opinVec{a}  = \argmin_{\bar{o}} \left\{ \selCost{a}(\bar{o}) + 
  \sum_{i \in \hood{a}} \frac{\wt{i}{a}(\wtSum{i}-\wt{i}{a})}{\wtSum{i}}\left(\bar{o}_i - \tilde{x}_i \right)^2
\right\}}. 
\end{equation}

\section*{Appendix B: Reconstruction results}

\label{app.reconstruction}

So that the reader can have a reasonable sense for the range of possible results, including
the nature of artefacts obtained using 
proactive message passing on memory factor networks, we include here images of 
the best and worst solutions for a variety of
problems and for each type of factor (memory table or subspace), 
first for reconstructing face images, and then reconstructing for music spectrograms.

\renewcommand{\arraystretch}{1.5}
\begin{figure}
	\centering
	\begin{tabular}{|m{0.15\textwidth}|m{0.15\textwidth}|m{0.15\textwidth}|m{0.15\textwidth}|m{0.15\textwidth}|}
		\hline
		Example & Original & Evidence & Memory Table Solution & Subspace Solution \\ \hline
		Best Memory Table &
		\includegraphics[scale=1.3]{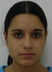}&
		\includegraphics[scale=1.3]{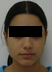} &
		\includegraphics[scale=1.3]{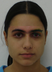} &
		\includegraphics[scale=1.3]{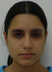} \\ 
		& & & MSE: 0.0036 & MSE: 0.0021 \\
		\hline
		Best Subspace &
		\includegraphics[scale=1.3]{graphics/eyes/best-orig-200a.png} &
		\includegraphics[scale=1.3]{graphics/eyes/best-start.png} &
		\includegraphics[scale=1.3]{graphics/eyes/best-mt.png}  &
		\includegraphics[scale=1.3]{graphics/eyes/best-ae.png} \\ 
		& & & MSE: 0.0036 & MSE: 0.0021 \\ \hline
		Worst Memory Table &
		\includegraphics[scale=1.3]{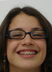} &
		\includegraphics[scale=1.3]{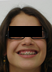} &
		\includegraphics[scale=1.3]{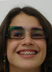} & 
		\includegraphics[scale=1.3]{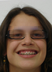} \\ 
		& & & MSE: 0.0346 & MSE: 0.0177 \\ \hline
		Worst Subspace &
		\includegraphics[scale=1.3]{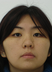} &
		\includegraphics[scale=1.3]{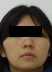} &
		\includegraphics[scale=1.3]{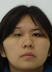} &
		\includegraphics[scale=1.3]{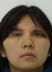} \\ 
		& & & MSE: 0.0138 & MSE: 0.0210 \\ \hline

	\end{tabular}
	\caption{Best and worst examples of eye reconstruction.}
	\label{fig:eye-extremes}
\end{figure}

\begin{figure}
	\centering
	\begin{tabular}{|m{0.15\textwidth}|m{0.15\textwidth}|m{0.15\textwidth}|m{0.15\textwidth}|m{0.15\textwidth}|}
		\hline
		Example & Original & Evidence & Memory Table Solution & Subspace Solution \\ \hline
		Best Memory Table &
		\includegraphics[scale=1.3]{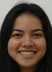} &
		\includegraphics[scale=1.3]{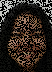} &
		\includegraphics[scale=1.3]{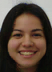} &
		\includegraphics[scale=1.3]{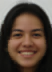} \\ 
		& & & MSE: 0.0023 & MSE: 0.0027 \\ \hline
		Best Subspace &
		\includegraphics[scale=1.3]{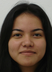} &
		\includegraphics[scale=1.3]{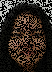} &
		\includegraphics[scale=1.3]{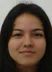} &
		\includegraphics[scale=1.3]{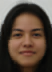} \\ 
		& & & MSE: 0.0034 & MSE: 0.0026 \\ \hline
		Worst Memory Table &
		\includegraphics[scale=1.3]{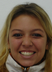} &
		\includegraphics[scale=1.3]{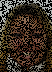} &
		\includegraphics[scale=1.3]{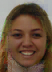} & 
		\includegraphics[scale=1.3]{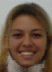} \\ 
		& & & MSE: 0.0111 & MSE: 0.0056 \\ \hline
		Worst Subspace &
		\includegraphics[scale=1.3]{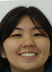} &
		\includegraphics[scale=1.3]{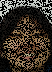} &
		\includegraphics[scale=1.3]{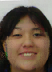} &
		\includegraphics[scale=1.3]{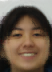} \\ 
		& & & MSE: 0.0086 & MSE: 0.0061 \\ \hline
	\end{tabular}
	\caption{Best and worst examples of reconstructing with randomly missing evidence.}
	\label{fig:drop-extremes}
\end{figure}

\begin{figure}
	\centering
	\begin{tabular}{|m{0.15\textwidth}|m{0.15\textwidth}|m{0.15\textwidth}|m{0.15\textwidth}|m{0.15\textwidth}|}
		\hline
		Example & Original & Evidence & Memory Table Solution & Subspace Solution \\ \hline
		Best Memory Table &
		\includegraphics[scale=1.3]{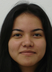} &
		\includegraphics[scale=1.3]{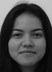} &
		\includegraphics[scale=1.3]{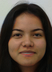} &
		\includegraphics[scale=1.3]{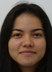} \\ 
		& & & MSE: 0.0005 & MSE: 0.0004 \\ \hline
		Best Subspace &
		\includegraphics[scale=1.3]{graphics/gray/best-orig-184a.png} &
		\includegraphics[scale=1.3]{graphics/gray/best-start.png} &
		\includegraphics[scale=1.3]{graphics/gray/best-mt.png} &
		\includegraphics[scale=1.3]{graphics/gray/best-ae.png} \\ 
		& & & MSE: 0.0005 & MSE: 0.0004 \\ \hline
		Worst Memory Table &
		\includegraphics[scale=1.3]{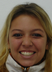} &
		\includegraphics[scale=1.3]{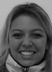} &
		\includegraphics[scale=1.3]{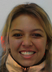} & 
		\includegraphics[scale=1.3]{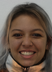} \\ 
		& & & MSE: 0.0058 & MSE: 0.0047 \\ \hline
		Worst Subspace &
		\includegraphics[scale=1.3]{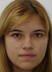} &
		\includegraphics[scale=1.3]{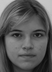} &
		\includegraphics[scale=1.3]{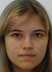} &
		\includegraphics[scale=1.3]{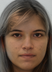} \\ 
		& & & MSE: 0.0030 & MSE: 0.0071 \\ \hline
	\end{tabular}
	\caption{Best and worst examples of colorization.}
	\label{fig:gray-extremes}
\end{figure}

\begin{figure}
	\centering
	\begin{tabular}{|m{0.15\textwidth}|m{0.15\textwidth}|m{0.15\textwidth}|m{0.15\textwidth}|m{0.15\textwidth}|}
		\hline
		Example & Original & Evidence & Memory Table Solution & Subspace Solution \\ \hline
		Best Memory Table &
		\includegraphics[scale=1.3]{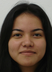} &
		\includegraphics[scale=1.3]{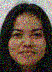} &
		\includegraphics[scale=1.3]{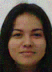} &
		\includegraphics[scale=1.3]{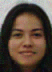} \\ 
		& & & MSE: 0.0035 & MSE: 0.0041 \\ \hline
		Best Subspace &
		\includegraphics[scale=1.3]{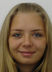} &
		\includegraphics[scale=1.3]{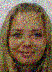} &
		\includegraphics[scale=1.3]{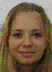} &
		\includegraphics[scale=1.3]{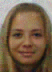} \\ 
		& & & MSE: 0.0081 & MSE: 0.0040 \\ \hline
		Worst Memory Table &
		\includegraphics[scale=1.3]{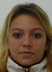} &
		\includegraphics[scale=1.3]{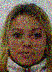} &
		\includegraphics[scale=1.3]{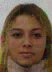} & 
		\includegraphics[scale=1.3]{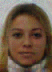} \\ 
		& & & MSE: 0.0106 & MSE: 0.0056 \\ \hline
		Worst Subspace &
		\includegraphics[scale=1.3]{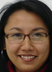} &
		\includegraphics[scale=1.3]{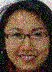} &
		\includegraphics[scale=1.3]{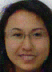} &
		\includegraphics[scale=1.3]{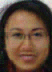} \\ 
		& & & MSE: 0.0073 & MSE: 0.0064 \\ \hline
	\end{tabular}
	\caption{Best and worst examples of reconstructing with noisy evidence.}
	\label{fig:noise-extremes}
\end{figure}


\begin{figure}
	\centering
	\begin{tabular}{|m{0.1\textwidth}|m{0.2\textwidth}|m{0.2\textwidth}|m{0.2\textwidth}|m{0.2\textwidth}|}
		\hline
		Example & Original & Evidence & Memory Table Solution & Subspace Solution \\ \hline
		Best Memory Table &
		\includegraphics[scale=0.25]{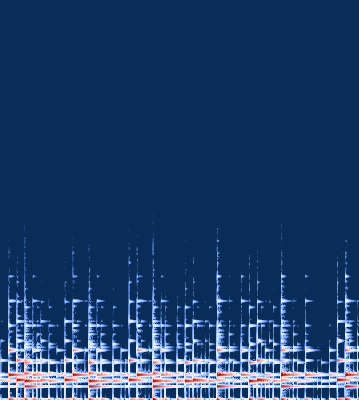} &
		\includegraphics[scale=0.25]{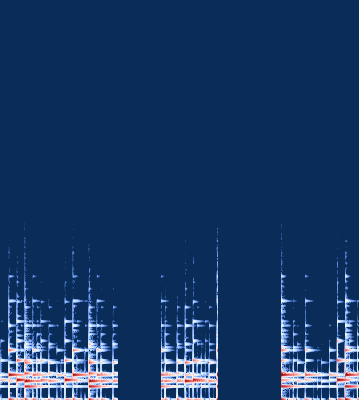} &
		\includegraphics[scale=0.25]{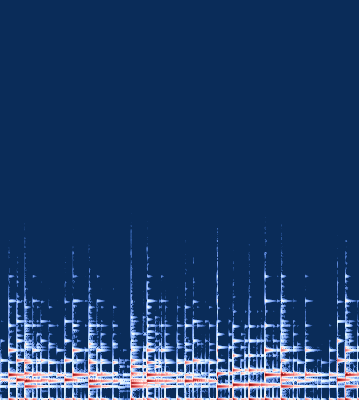} &
		\includegraphics[scale=0.25]{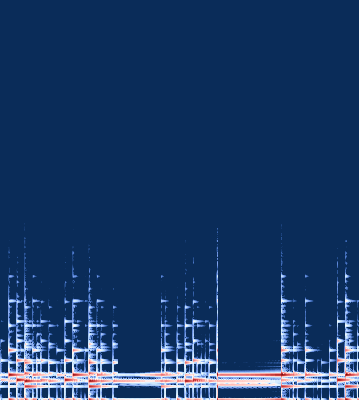} \\ 
		& & & MSE: 8.205 & MSE: 5.073 \\ \hline
		Best Subspace &
		\includegraphics[scale=0.25]{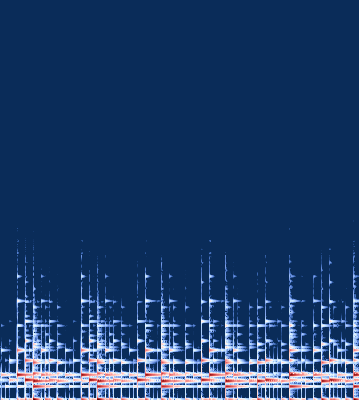} &
		\includegraphics[scale=0.25]{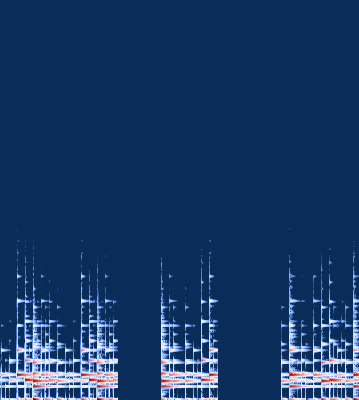} &
		\includegraphics[scale=0.25]{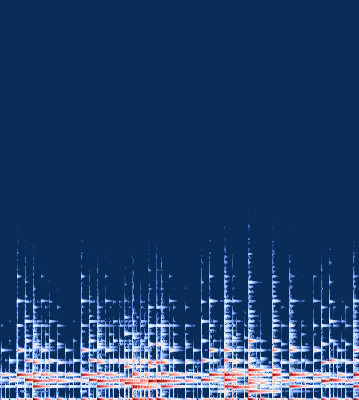} &
		\includegraphics[scale=0.25]{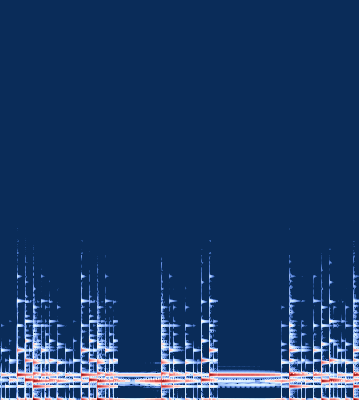} \\ 
		& & & MSE: 10.341 & MSE: 3.712 \\ \hline
		Worst Memory Table &
		\includegraphics[scale=0.25]{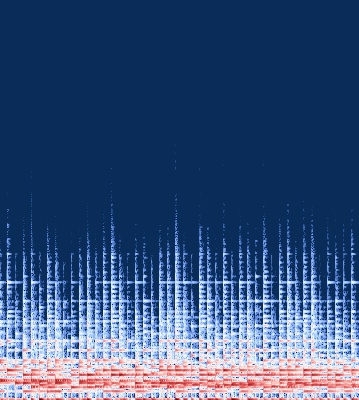} &
		\includegraphics[scale=0.25]{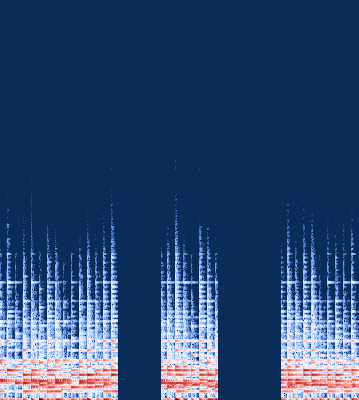} &
		\includegraphics[scale=0.25]{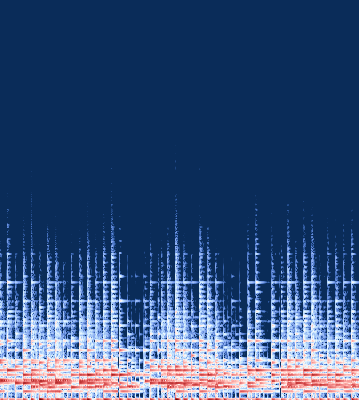} & 
		\includegraphics[scale=0.25]{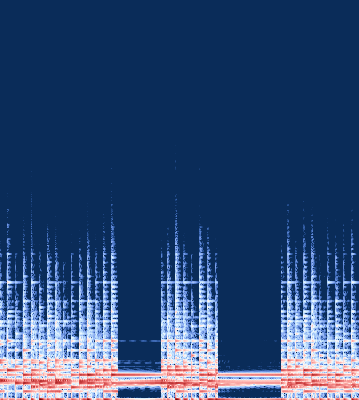} \\ 
		& & & MSE: 28.042 & MSE: 20.366 \\ \hline
		Worst Subspace &
		\includegraphics[scale=0.25]{graphics/music-missing/worst-orig-6a.png} &
		\includegraphics[scale=0.25]{graphics/music-missing/worst-start.png} &
		\includegraphics[scale=0.25]{graphics/music-missing/worst-mt.png} & 
		\includegraphics[scale=0.25]{graphics/music-missing/worst-ae.png} \\ 
		& & & MSE: 28.042 & MSE: 20.366 \\ \hline
	\end{tabular}
	\caption{Best and worst examples of reconstructing music with missing sections.}
	\label{fig:music-missing-extremes}
\end{figure}

\begin{figure}
	\centering
	\begin{tabular}{|m{0.1\textwidth}|m{0.2\textwidth}|m{0.2\textwidth}|m{0.2\textwidth}|m{0.2\textwidth}|}
		\hline
		Example & Original & Evidence & Memory Table Solution & Subspace Solution \\ \hline
		Best Memory Table &
		\includegraphics[scale=0.25]{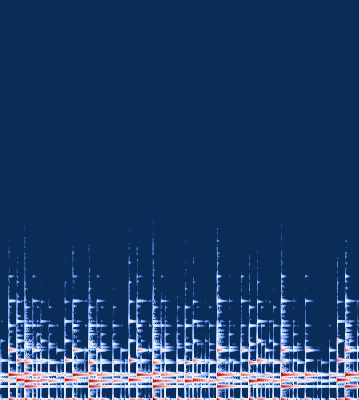} &
		\includegraphics[scale=0.25]{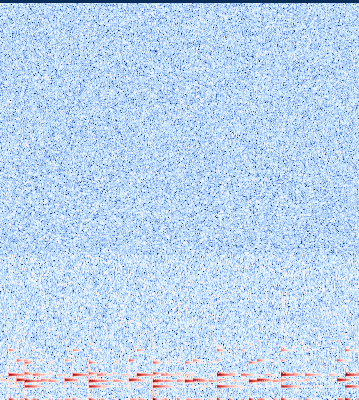} &
		\includegraphics[scale=0.25]{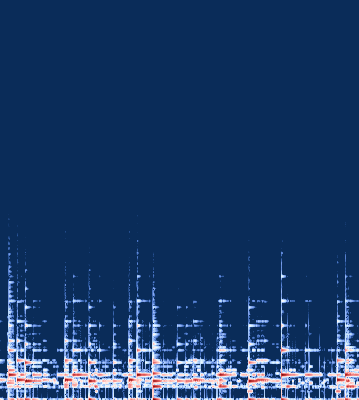} &
		\includegraphics[scale=0.25]{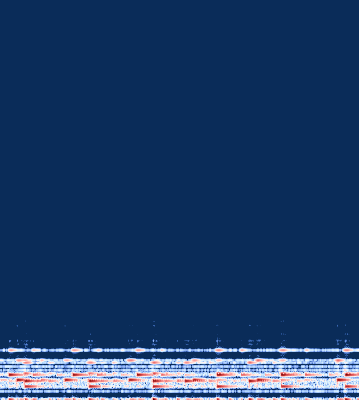} \\ 
		& & & MSE: 4.932 & MSE: 0.635 \\ \hline
		Best Subspace &
		\includegraphics[scale=0.25]{graphics/music-noise/best-orig-10b.png} &
		\includegraphics[scale=0.25]{graphics/music-noise/best-start.png} &
		\includegraphics[scale=0.25]{graphics/music-noise/best-mt.png} &
		\includegraphics[scale=0.25]{graphics/music-noise/best-ae.png} \\ 
		& & & MSE: 4.932 & MSE: 0.635 \\ \hline
		Worst Memory Table &
		\includegraphics[scale=0.25]{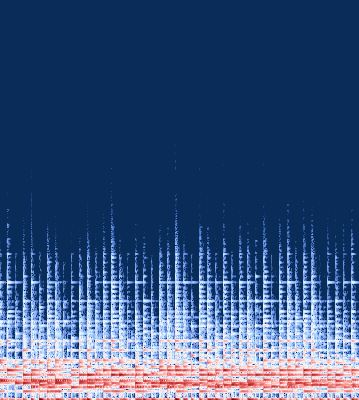} &
		\includegraphics[scale=0.25]{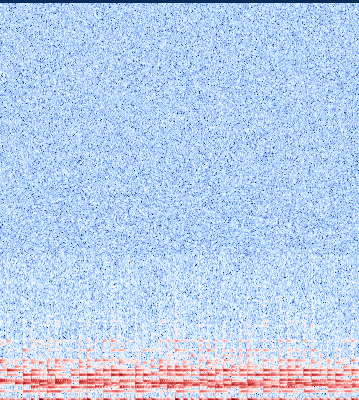} &
		\includegraphics[scale=0.25]{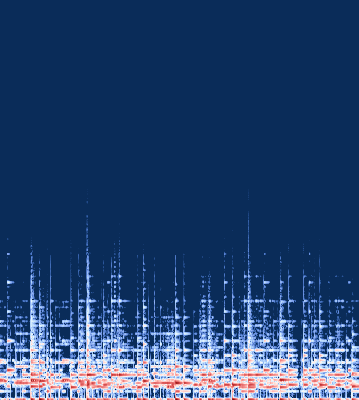} & 
		\includegraphics[scale=0.25]{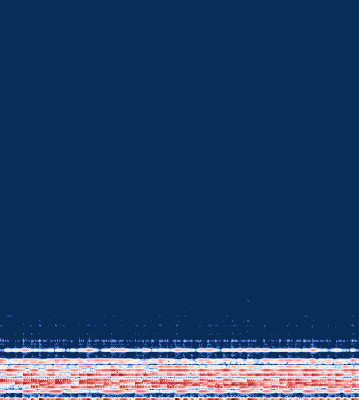} \\ 
		& & & MSE: 42.328 & MSE: 4.304 \\ \hline
		Worst Subspace &
		\includegraphics[scale=0.25]{graphics/music-noise/worst-orig-6a.png} &
		\includegraphics[scale=0.25]{graphics/music-noise/worst-start.png} &
		\includegraphics[scale=0.25]{graphics/music-noise/worst-mt.png} & 
		\includegraphics[scale=0.25]{graphics/music-noise/worst-ae.png} \\ 
		& & & MSE: 42.328 & MSE: 4.304 \\ \hline
	\end{tabular}
	\caption{Best and worst examples of reconstructing music with noise added.}
	\label{fig:music-noise-extremes}
\end{figure}

\bibliography{main} 

\end{document}